\pdfoutput=1
\documentclass[conference]{IEEEtran}
\usepackage{times}
\usepackage{graphicx} 
\usepackage{epsfig} 
\usepackage{times} 
\usepackage{amsmath} 
\usepackage{amssymb}  
\usepackage{subcaption}
\usepackage{mathtools}
\let\labelindent\relax 
\usepackage{enumitem}
\usepackage[font={small}]{caption}
\usepackage{mathrsfs}
\usepackage{amsfonts}
\usepackage{algorithm}
\usepackage{algorithmic}
\usepackage[short]{optidef}
\usepackage{balance}
\usepackage{braket}
\usepackage{pdfpages}
\usepackage{color}
\usepackage[square,sort,comma,numbers]{natbib}
\usepackage{multicol}
\usepackage{bm}
\usepackage{tikz}
\usepackage[bookmarks=true]{hyperref}
\usepackage{multirow}
\usepackage{etoolbox,lipsum}

\newtheorem{remark}{Remark}

\newcommand{\insertfig}{%
  \makebox[0pt]{\includegraphics[width=\linewidth]{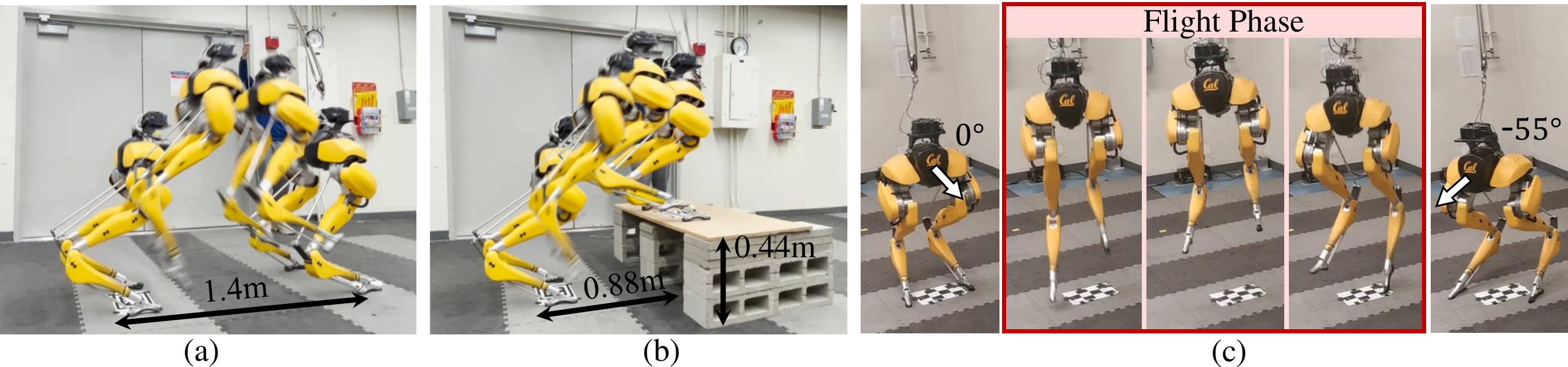}\label{fig:intro}} \\
  \small{Fig.~1: Representative dynamic jumping maneuvers performed by a bipedal robot Cassie using the proposed goal-conditioned control policies. From left to right: (a) the robot jumps over $1.4$ m and lands at the given target; (b) the robot jumps to a target that is $0.88$ m in front of the robot and $0.44$ m above the ground, and (c) the robot jumps in place while turning $55^\circ$ with a command to turn $60^\circ$ in place. The policies are trained in simulation and deployed on the hardware without further tuning. Video is at: \url{https://youtu.be/aAPSZ2QFB-E}.
  }
}

\makeatletter
\apptocmd{\@maketitle}{~~~~~~~~~~~~~~~~~~~~~~~~~~~~~~~~~~~~~~~~~~~~~~~~~~~~~~~~~~~~~~~~~~~~~~~~~~\insertfig}{}{} 
\makeatother

\begin{document}

\title{Robust and Versatile Bipedal Jumping Control through Reinforcement Learning}

\author{Zhongyu Li$^1$, Xue Bin Peng$^2$, Pieter Abbeel$^1$, Sergey Levine$^1$, Glen Berseth$^{3,4}$, and Koushil Sreenath$^1$\\
\normalsize{$^1$University of California, Berkeley,~~~~$^2$Simon Fraser University,~~~~$^3$Universit\'e de Montr\'eal,~~~~$^4$Mila}\\
\normalsize{Email: zhongyu\_li@berkeley.edu, xbpeng@sfu.ca, pabbeel@cs.berkeley.edu, svlevine@eecs.berkeley.edu,} \\ \normalsize{glen.berseth@mila.quebec, koushils@berkeley.edu}}

\maketitle

\begin{abstract}
This work aims to push the limits of agility for bipedal robots by enabling a torque-controlled bipedal robot to perform robust and versatile dynamic jumps in the real world.
We present a reinforcement learning framework for training a robot to accomplish a large variety of jumping tasks, such as jumping to different locations and directions.   
To improve performance on these challenging tasks, we develop a new policy structure that encodes the robot's long-term input/output (I/O) history while also providing direct access to a short-term I/O history. 
In order to train a versatile jumping policy, we utilize a multi-stage training scheme that includes different training stages for different objectives. 
After multi-stage training, the policy can be directly transferred to a real bipedal Cassie robot.
Training on different tasks and exploring more diverse scenarios lead to highly robust policies that can exploit the diverse set of learned maneuvers to recover from perturbations or poor landings during real-world deployment. 
Such robustness in the proposed policy enables Cassie to succeed in completing a variety of challenging jump tasks in the real world, such as standing long jumps, jumping onto elevated platforms, and multi-axes jumps. 

\end{abstract}

\IEEEpeerreviewmaketitle

\section{Introduction}
One question that has been lingering since the first creation of bipedal robots is: how can we enable such complex robots to traverse complex environments using agile and robust maneuvers~\cite{hirose2007honda,Raibert-1984-15614}? For example, creating agile controllers that enable bipedal robots to jump over a given distance or onto different elevations can enable greater mobility in unstructured environments.
However, jumping is a challenging skill to control for bipedal robots.
During a standing jump, the robot needs to push its body off the ground and break contact, leap into a flight phase where the robot is underactuated,
and then make contact again when it lands on its legs. During landing, the robot needs to not only recover from the large impact impulse, but also stick the landing and remain standing. 
All of these events occur within a very short amount of time (typically less than 2 seconds).
These events lead to a hybrid system that switches between modes with different contact configurations (\textit{e.g.}, taking-off, flight, and landing).
Planning and controlling such discontinuous dynamics, especially for a high-dimensional, nonlinear, and underactuated bipedal robot present a very challenging task~\cite{posa2014direct}. 
This challenge is further compounded when the robot needs to accurately land on a given target, where the robot will have to produce the precise translational and angular momentum at take-off in order to land at the desired location~\cite{yim2018precision}.

\begin{table*}[!t]
\centering
\caption{Benchmark with previous work tackling jumping control using optimal control (OC) or reinforcement learning (RL) on the bipedal robot Cassie in the real world.}
\label{tab:related}
\begin{tabular}{cccccccccc}
\hline
\multirow{2}{*}{} &
  \textbf{Controlled} &
  \textbf{Apex} &
  \textbf{Longest} &
  \multicolumn{5}{c}{\textbf{Maximum Leap Distance}} \\ \cline{5-9} 
   &
  \textbf{Landing Pose} &
  \textbf{Foot Clearence} &
  \textbf{Flight Phase} &
  \textbf{Forward} &
  \textbf{Backward} &
  \textbf{Lateral} &
  \textbf{Turning} &
  \textbf{Elevation} \\ \hline
Aperiodic Hop by OC~\cite{xiong2018bipedal}  & No & 0.18m        & 0.42s  & $\sim$0.5m* & 0m & 0m & 0$^\circ$ & 0m                \\
Aperiodic Hop by OC~\cite{yang2021impact}    & No & 0.15m        & 0.33s*  & $\sim$0.3m* & 0m & 0m & 0$^\circ$ & 0m \\
Periodic Hop by RL~\cite{siekmann2021sim}   & No & $\sim$0.16m* & 0.33s* & $\sim$0.5m* & 0m & 0m & 0$^\circ$ & $\sim$0.15m*      \\ \hline
\textbf{Ours} (Aperiodic Jump by RL) &
  \textbf{Yes} &
  \textbf{0.47m} &
  \textbf{0.58s} &
  \textbf{1.4m} &
  \textbf{-0.3m} &
  \textbf{$\pm$0.3m} &
  \textbf{$\pm$55$^\circ$} &
  \textbf{0.44m} \\ \hline
\end{tabular}\\
\raggedright
~~*Not provided in the paper and the listed value is roughly estimated based on the comparison with the background environment in the accompanying video.
\vskip-10pt
\end{table*}

Model-based optimal control (OC) frameworks have made notable progress in controlling bipedal robots, including jumping, but they depend on carefully crafted models of the robot and the complex contact dynamics~\cite{koditschek1991analysis,seyfarth2003swing,manchester2020variational}. These methods typically require manual design of task-specific control structures \cite{goswami2009planar,chen2020underactuated,zhang2020biologically}, and simplified dynamics models, which provides only a coarse approximation of the robot's full-order dynamics. Furthermore, pre-defined or pre-computed contact sequences are often needed to reduce online optimization complexity~\cite{park2015online,xiong2018bipedal,nguyen2019optimized}. Such limitations have restricted previous model-based methods to only performing a fixed hop on torque-controlled bipedal robots like Cassie~\cite{xiong2018bipedal,yang2021impact}.
By leveraging the robot's full-order dynamics, model-free reinforcement learning~(RL) has shown some success in highly-dynamic locomotion control on quadrupedal robots~\cite{peng2020learning,margolis2021learning,ji2022concurrent}.
However, compared to quadrupeds that are inherently more stable, RL-based methods still struggle when applied to bipedal robots for more dynamic and aggressive maneuvers in the real world.   
Given that the most relevant prior RL-based system is only able to perform periodic hopping on Cassie~\cite{siekmann2021sim}, it remains an open question how more dynamic bipedal skills can be achieved in the real world, such as standing long jumps that could be more challenging than periodic motions~\cite[Sec.~I]{goswami2009planar}.

\subsection{Objective of this Paper}
In this paper, our objective is to explore the possibility of creating a robust and versatile jumping controller for a bipedal robot that enables it to land at different target locations, as shown in Fig.~1, and validating the advantages brought by learning with different maneuvers using reinforcement learning. 
We refer a \emph{skill} to a kind of legged locomotion method, such as walking, jumping, or running.
We further denote a \emph{task} as using a locomotion skill to accomplish a \emph{goal}. For instance, one task could be the task of walking (skill) at a desired speed (goal), or as in this work, achieving the task of jumping (skill) to a target configuration (goal).
We use the term \emph{goal-conditioned} to refer to a policy that can perform a variety of jumping tasks, such as jumping over various desired distances and/or directions, conditioned on the given goal.
We hypothesize that, by exploring different jumping tasks, a versatile jumping controller can be more robust as it allows the robot to leverage more diverse learned tasks to maintain stability during dynamic maneuvers.
For example, in order to recover from a large ground impulse on landing, the robot can quickly switch to another learned task, such as a hop to a different location, which can allow for a more graceful recovery than simply continuing with the original behavior.

\subsection{Contributions}
The core contribution of this work is the the development of the first system that enables a life-sized torque-controlled bipedal robot to perform \emph{versatile jumping maneuvers with controlled landing locations} in the real world. 
The robot is first trained in simulation with reinforcement learning and domain randomization.
Training does not require an explicit contact sequence, and the learning algorithm automatically develops different contact sequences for different jumping goals.
In order to successfully transfer the learned skill for such dynamic maneuvers from simulation to the real world, we utilize two new design decisions. 
First, we present a new policy structure that encodes the robot's long-term Input-Output (I/O) history while also providing direct access to a short-term I/O history.
By training the model in an end-to-end manner, we show that such a structure can outperform previously proposed architectures (Fig.~\ref{fig:learning_logs}).
Second, we demonstrate that training the controller for a diverse range of goals improves the robustness of the controller by using the maneuvers learned from different jumping tasks to recover from unstable states. 
We show that this robustness cannot be easily obtained by training only for a single jumping goal (Fig.~\ref{fig:bm_single_sim}).  
Combining these two techniques, we enable the bipedal Cassie robot to perform (1) several aggressive jumps, such as a long jump (1.4m ahead) and a high jump onto a 0.44m-tall platform (Fig.~1), (2) various agile multi-axes bipedal jumps (Fig.~\ref{fig:snapshot_timeline}, Fig.~\ref{fig:snapshot_rest}), and (3) to utilize diverse learned jumping maneuvers to recover from external perturbations or impacts (Fig.~\ref{fig:snapshot_rest}), in the real world. 
We hope this paper can serve as a step forward for enabling more diverse, dynamic, and robust legged locomotion skills.

\section{Related Work}
Prior work tackling dynamic locomotion skills such as jumping with legged robots can be broadly categorized as corresponding to model-based optimal control (OC) and model-free reinforcement learning (RL). 
Table~\ref{tab:related} compares our work with the most related prior efforts on the bipedal robot Cassie.

\subsubsection{Model-based optimal control for legged jumping}
Prior model-based methods for legged jumping control usually build up a layered optimization scheme, which includes offline trajectory optimization with detailed models of the robot's dynamics and ground contacts~\cite{dai2014whole,ding2020kinodynamic,nguyen2019optimized,song2022optimal}, and online controllers that leverage simplified models of the robot's dynamics~\cite{rutschmann2012nonlinear,ugurlu2012hopping,yang2021impact,nguyen2022continuous}. 
In order to optimize trajectories for jumping, which needs to switch among modes with different underlying dynamics, there are two commonly employed solutions: (i) relying on human-specified contact sequences~\cite{park2015online,xiong2018bipedal,katz2019mini,chen2020underactuated,gilroy2021autonomous}, which is not scalable to different jump distances and/or directions, or (ii) leveraging contact-implicit optimization~\cite{posa2014direct,chatzinikolaidis2020contact,zhu2021contact,drnach2021robust,landry2022bilevel}
which plans through contacts to avoid breaking the trajectory or using computationally expensive mixed-integer programming~\cite{ding2018single,ding2020kinodynamic,aceituno2017simultaneous}. 
However, due to the computational challenges of optimization, both of the above-mentioned methods are still limited to offline computation for legged robots.
As we show in this work, by training with different jumping goals offline, Cassie can automatically generate the appropriate contact sequences during online execution for achieving robust jumping.

In the case of controllers for aperiodic dynamic jumps, many previous efforts required a separate landing controller to stabilize the robot from the large landing impacts~\cite{goswami2009planar,xiong2018bipedal,nguyen2019optimized,jeon2022online}.
However, this approach usually requires a contact estimator and needs to fuse the noisy robot proprioceptive measurements to estimate the contact states, which on its own could be a challenging problem~\cite{pfrommer2021contactnets,hartley2020contact,lin2021legged}.
Furthermore, while there are a few prior attempts addressing \textit{precise} jumping control on low-dimensional single-legged robots~\cite{yim2018precision,vatavuk2021precise} and a quadrupedal robot~\cite{nguyen2022continuous}, mostly in simulation~\cite{vatavuk2021precise,nguyen2022continuous}, most prior work on bipedal jumping only focus on  single vertical jumps with the landing location not controlled~\cite{goswami2009planar,ugurlu2012hopping,xiong2018bipedal,yang2021impact}. 
In this work, the proposed jumping controller demonstrates the capacity to control the landing pose of the bipedal robot without position feedback or explicit contact estimation.

\subsubsection{Model-free RL for legged locomotion control}
In recent years, we have seen exciting progress in using deep RL to learn locomotion controllers for quadrupedal robots~\cite{lee2020learning,feng2022genloco, margolisyang2022rapid,bellegarda2022robust} and bipedal robots~\cite{yu2019sim,xie2020learning,siekmann2020learning,li2021reinforcement,rodriguez2021deepwalk,castillo2021robust} in the real world. 
Since it is challenging in general to learn a single policy with RL to perform various tasks~\cite{kalashnikov2021mt}, many prior works focus on learning a single-goal policy~\cite{peng2020learning,margolis2021learning,yu2022dynamic,bogdanovic2022model} for legged robots, such as just forward walking at a constant speed~\cite{xie2018feedback,kumar2021rma,feng2022genloco}. 
There have been efforts to obtain more versatile policies, such as walking at different velocities using different gaits, while following different commands~\cite{fu2021minimizing,rodriguez2021deepwalk,li2022human,fu2022deep}, which requires more extensive tuning due to the lack of a gait prior.
Providing the robot with different reference motions for different goals can be helpful, but requires additional parameterization of the reference motions (\textit{e.g.}, a gait library)~\cite{li2021reinforcement,ji2022hierarchical,huang2022creating,batke2022optimizing}, policy distillation~\cite{xie2020learning}, or a motion prior~\cite{peng2021amp,escontrela2022adversarial,vollenweider2022advanced}. 
There is also a line of research to explicitly provide contact sequences for legged robots~\cite{siekmann2021sim,shao2021learning,margolis2022walk,bellegarda2022cpg}. 
However, such methods are prescriptive and provide little opportunity for the robot to deviate from the contact plan, limiting the flexibility with which it can respond to perturbations.
In this work, we show that a versatile policy can enhance the robustness of a jumping policy by intelligently employing a variety of learned tasks to react to perturbations.

\subsubsection{Sim-to-real transfer for legged robots}
To tackle sim-to-real transfer for RL-based methods, some works have sought to directly train policies directly in the real world~\cite{haarnoja2018learning,wu2022daydreamer,smith2022walk}, but most of the prior work, especially for dynamic skills, leverages a simulator to train the legged robot with extensive dynamics randomization~\cite{peng2018sim} and then zero-shot transfer to the real world~\cite{li2021reinforcement,siekmann2021sim,lee2020learning,kumar2021rma,feng2022genloco} or finetune with real-world data~\cite{peng2020learning,smith2022legged,ji2022hierarchical}. 
Since performing rollout on the hardware of human-scale bipedal robots is expensive, we use the zero-shot transfer method. 
In order to realize this, there are two widely-adopted techniques: (i) end-to-end training a policy by providing the robot with a proprioceptive short-term history~\cite{li2021reinforcement,feng2022genloco,huang2022creating} or long-term history~\cite{peng2018sim,siekmann2020learning,shao2021learning}, (ii) teacher-student training that first obtains a teacher policy with privileged information of the environment by RL, then uses this policy to supervise the training of a student policy that only has access of onboard-available observations~\cite{yu2019sim,lee2020learning,kumar2021rma,ji2022concurrent,margolisyang2022rapid,fu2022deep}, which shows advantages over the end-to-end training method~\cite{kumar2021rma,kumar2022adapting}. 
However, here we show that, for the dynamic control of bipedal robots, by training the robot in an end-to-end way with a newly-proposed policy structure, we can realize a better learning performance over the teacher-student method which separates the training process and results in increased training time and data.

\section{Background and Preliminaries}
In this section, we provide a brief introduction to our experimental platform, Cassie and the background of goal-conditioned reinforcement learning. 

\subsection{Floating-base Model of Cassie}~\label{subsec:cassie_model}
We use Cassie as the experimental platform in this work. Cassie (see Fig.~1) is a life-sized bipedal robot and is around $1.1$ meter tall, with a weight of $31$ Kg.
It is a dynamic and underactuated system, with $5$ actuated motors (abduction $q_1$, rotation $q_2$, thigh $q_3$, knee $q_4$, and toe $q_7$) and $2$ passive joints (shin $q_5$ and tarsus $q_6$) connected by leaf springs on its Left and Right leg. 
We denote the motor positions as $\mathbf{q}_m=[q^{L/R}_{1,2,3,4,7}] \in \mathbb{R}^{10}$.
The 6 degree of freedom (DoF) floating base (pelvis) can be represented with translational positions (sagittal $q_x$, lateral $q_y$, vertical $q_z$) and rotational positions (roll $q_\psi$, pitch $q_\theta$, and yaw $q_\phi$). 
In total, the robot has $20$ DoFs $\mathbf{q} \in \mathbb{R}^{20}$. 
For more details about Cassie's configuration, we refer readers to~\cite[Fig.~2]{xiong2018bipedal}. 
The observable joint positions on Cassie are denoted as $\mathbf{q}^o =[q_{\psi,\theta,\phi}, \mathbf{q}_m, \dot{q}_{x,y,z}, \dot{\mathbf{q}}_m] \in \mathbb{R}^{26}$, which can be obtained from onboard joint encoders and IMUs, while the base linear velocity $\dot{q}_{x,y,z}$ can be estimated with an EKF~\cite{xie2020learning}.

\subsection{RL Background and Goal-Conditioned Policy}~\label{subsec:task}
We formulate the locomotion control problem as a Markov decision process (MDP). 
At each timestep $t$, the agent (\textit{i.e.}, the robot) observes the environment state $\mathbf{s}_t$, and the policy $\pi$ produces a distribution over the actions, $\pi(\mathbf{a}_t|\mathbf{s}_t)$, conditioned on the state. 
The agent then executes the action $\mathbf{a}_t$ sampled from the policy, interacts with the environment, makes an observation of the environment's new states $\mathbf{s}_{t+1}$, and receives a reward $r_t$. 
The objective of RL is to maximize the expected accumulative reward (return) the agent received over the course of an episode $\mathbb{E}[\Sigma^{T}_{t=0} \gamma^t r_t]$ where $\gamma$ is a discount factor and $T$ is the episode length. 
In order to obtain a policy that can accomplish different goals, we provide a goal $\mathbf{c}$ which parameterizes the task and the policy $\pi(\mathbf{a}_t|\mathbf{s}_t, \mathbf{c})$ is then also conditioned on the given goal $\mathbf{c}$ to perform different tasks.  

\textit{Task Parameterization:} In our jumping task, the goal $\mathbf{c}$ specifies target commands for a desired jump $\mathbf{c}=[c_x,c_y,c_z,c_\phi]$, which consists of the target location $c_{x,y}$ on the horizontal plane, elevation $c_z$ in the vertical direction, and turning direction $c_{\phi}$ after the agent lands, calculated based on the robot's pose before the jump, \textit{i.e.}, in the local frame of robot's starting pose.
Please note that the change in elevation $c_z$ is defined as the change of the floor height, instead of the change of the robot's base height.


\section{Multi-Stage Training for Versatile Jumps}~\label{sec:methodology}

We now describe our multi-stage training framework for developing goal-conditioned jumping policies. The training environment is developed in a simulation of Cassie using MuJoCo~\cite{todorov2012mujoco,cassiemj}.

\subsection{Overview of the Multi-Stage Training Schematic}~\label{sec:overview}

\begin{figure}
\centering
\includegraphics[width=\linewidth]{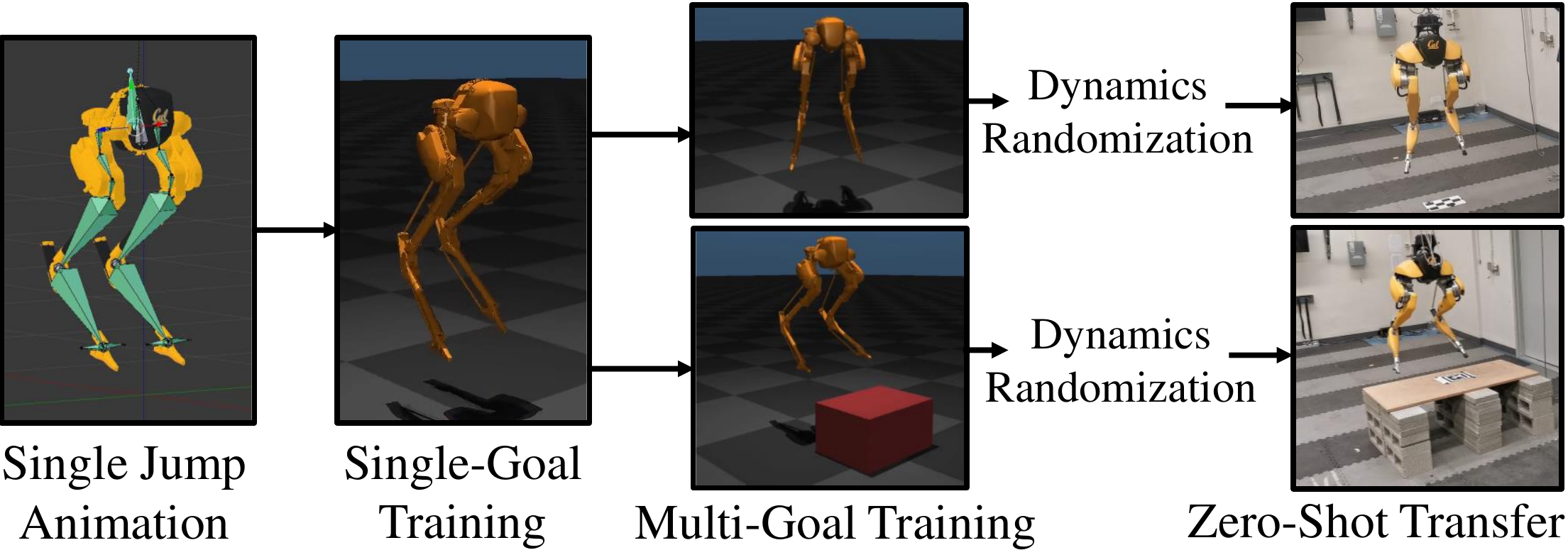}
\refstepcounter{figure}
\caption{The schematic to train the robot to perform versatile jumping skills in the real world starting with a reference motion of a single jumping animation. This framework consists of three stages. In the first stage, we focus on training the robot to imitate the animation while performing a single jump from scratch. After the robot is good at achieving the single goal, we randomize the goal (to land at different locations and different turning directions/elevations) that is assigned to the robot during each training episode. After these two stages of training, we extensively randomize the dynamics properties of the environment in simulation in order to improve the robustness of the robot during the zero-shot transfer from sim to real.}
\label{fig:pipeline}
\vskip-10pt
\end{figure}

Our goal is to develop a locomotion control policy for jumping skills that can perform targeted jumps to different locations. 
However, due to the challenging nature of jumping, it can be difficult to directly train a policy to perform a large variety of jumps from scratch. We observed that training policies from scratch to perform a large variety of jumps tends to lead to the robot adopting very conservative behaviors or even failing to learn to jump.  
Therefore, we use a multi-stage training scheme that consists of 3 stages, as illustrated in Fig.~\ref{fig:pipeline}: (1) single-goal training, (2) multi-goal training, and (3) dynamics randomization. All stages of training are performed in simulation, but as we show in our experiments, the resulting models can then be directly deployed on a real Cassie robot.
In \textit{Stage 1}, the model is trained on a single goal $\mathbf{c}$, \textit{i.e.}, jumping in place.   
This stage of training results in a policy that is trained from scratch to specialize in a single task in simulation.
Next, in \textit{Stage 2}, the goal $\mathbf{c}$ is randomized every episode to train the robot to jump to different targets. 
In this stage, the focus is primarily on performing the commanded task in the simulated environment. 
Finally, in \textit{Stage 3}, we introduce extensive domain randomization on the simulation environment, while also randomizing the goal, in order to improve the robustness and generalization of the policy for sim-to-real transfer. 
At each stage, the reward and episode designs of the MDP may vary in order to produce more effective policies for the objectives of the given stage. 

In the rest of this section, we focus on explaining the details of Stages 2\&3 in training, which share the same reward and episode design and cover both multi-goal training and domain randomization. Stage 1 training, which is different in the choice of hyperparameters, is detailed in Appendix~\ref{sec:training_stage1}.

\subsection{Reference Motion}
To initialize the training process, we provide a \textit{single} jumping reference motion. 
The reference motion is a human-authored animation of Cassie jumping in place, created in a $3D$ creation suite~\cite{li2020animated}, as presented in Fig.~\ref{fig:pipeline}.
This animated jump has an apex foot height of $0.5$~m and an apex pelvis height of $1.1$~m and has a timespan $T_J$ of $1.66$ second (\textit{i.e.}, $T_J=1.66$).
This reference motion is only a kinematically-feasible motion for the agent and is not optimized to be dynamically feasible.
After the end of the jumping animation, the reference motion will be set to a fixed standing pose for the robot.

\subsection{Reward}
The design of the reward function is important to encourage the robot to jump with agility. 
We further spilt the reward within a stage into two phases: before landing ($t\leq T_J$) and after ($t > T_J$), and the reward needs to vary based on these phases because the desired robot's behavior is different: performing aggressive jump versus stationary standing.   

We here define a function: 
\begin{equation}
    r(u,v) = \exp(-\alpha||u-v||_2^2)
\end{equation}
where $r(u,v)\in (0,1]$ defines a reward component that encourage the two vector $u$ and $v$ to be as close as possible, scaled by $\alpha>0$ that balance the units.
The reward $r_t$ the agent receives at each timstep is a weighted summation of different components, $r_t=(\mathbf{w}/||\mathbf{w}||_1)^T\mathbf{r} \in [0,1]$.
The component vector $\mathbf{r}$ and weight vector $\mathbf{w}$ are detailed in Table~\ref{tab:reward}.
The reward used here consists of three main components: reference motion tracking, task completion, and smoothing term.


\begin{table}[t]
\centering
\scriptsize
\caption{A list of components of reward $r_t$ which is a weighted summation of the listed items. The weight of each term is scheduled based on the jumping phase and training stage.}
\label{tab:reward}
\begin{tabular}{|ccccc|}
\hline
\multicolumn{1}{|c|}{\multirow{3}{*}{\textbf{Reward Component} $\mathbf{r}$}} &
  \multicolumn{4}{c|}{\textbf{Weight} $\mathbf{w}$} \\ \cline{2-5} 
\multicolumn{1}{|c|}{} &
  \multicolumn{2}{c|}{\textbf{Stage 1}} &
  \multicolumn{2}{c|}{\textbf{Stage 2, 3}} \\ \cline{2-5} 
\multicolumn{1}{|c|}{} &
  \multicolumn{1}{c|}{$t\leq T_{J}$} &
  \multicolumn{1}{c|}{$t>T_{J}$} &
  \multicolumn{1}{c|}{$t\leq T_{J}$} &
  $t>T_{J}$ \\ \hline
\multicolumn{5}{|c|}{\textbf{Reference Motion Tracking}} \\ \hline
\multicolumn{1}{|c|}{Motion position: $r(\mathbf{q}_m, \mathbf{q}^r_m(t))$} &
  \multicolumn{1}{c|}{15} &
  \multicolumn{1}{c|}{15} &
  \multicolumn{1}{c|}{7.5} &
  15 \\ \hline
\multicolumn{1}{|c|}{Pelvis height: $r(q_z, q^r_z(t)+c_z)$} &
  \multicolumn{1}{c|}{5} &
  \multicolumn{1}{c|}{5} &
  \multicolumn{1}{c|}{3} &
  3 \\ \hline
\multicolumn{1}{|c|}{Foot height: $r(e_z, e^r_z(t)+c_z)$} &
  \multicolumn{1}{c|}{10} &
  \multicolumn{1}{c|}{10} &
  \multicolumn{1}{c|}{10} &
  10 \\ \hline
\multicolumn{5}{|c|}{\textbf{Task Completion}} \\ \hline
\multicolumn{1}{|c|}{Pelvis position: $r(q_{x,y}, c_{x,y})$} &
  \multicolumn{1}{c|}{12.5} &
  \multicolumn{1}{c|}{12.5} &
  \multicolumn{1}{c|}{15} &
  15 \\ \hline
\multicolumn{1}{|c|}{Pelvis velocity: $r(\dot{q}_{x,y}, \dot{q}^d_{x,y})$} &
  \multicolumn{1}{c|}{0} &
  \multicolumn{1}{c|}{3} &
  \multicolumn{1}{c|}{12.5} &
  12.5 \\ \hline
\multicolumn{1}{|c|}{Orientation: $r(q_{\psi,\theta,\phi}, [0,0,c_\phi])$} &
  \multicolumn{1}{c|}{12.5} &
  \multicolumn{1}{c|}{12.5} &
  \multicolumn{1}{c|}{10} &
  10 \\ \hline
\multicolumn{1}{|c|}{Angular rate: $r(q_{\psi,\theta,\phi}, [0,0,\dot{q}^d_\phi])$} &
  \multicolumn{1}{c|}{3} &
  \multicolumn{1}{c|}{3} &
  \multicolumn{1}{c|}{10} &
  10 \\ \hline
\multicolumn{5}{|c|}{\textbf{Smoothing}} \\ \hline
\multicolumn{1}{|c|}{Ground Impact: $r(F_z, 0)$} &
  \multicolumn{1}{c|}{5} &
  \multicolumn{1}{c|}{0} &
  \multicolumn{1}{c|}{10} &
  0 \\ \hline
\multicolumn{1}{|c|}{Torque Consumption: $r(\bm{\tau}, 0)$} &
  \multicolumn{1}{c|}{3} &
  \multicolumn{1}{c|}{3} &
  \multicolumn{1}{c|}{3} &
  15 \\ \hline
\multicolumn{1}{|c|}{Motor velocity: $r(\dot{\mathbf{q}}_m, 0)$} &
  \multicolumn{1}{c|}{0} &
  \multicolumn{1}{c|}{15} &
  \multicolumn{1}{c|}{0} &
  25 \\ \hline  
\multicolumn{1}{|c|}{Joint acceleration: $r(\ddot{\mathbf{q}}, 0)$} &
  \multicolumn{1}{c|}{3} &
  \multicolumn{1}{c|}{10} &
  \multicolumn{1}{c|}{0} &
  5 \\ \hline
\multicolumn{1}{|c|}{Change of action: $r(\mathbf{a}_t, \mathbf{a}_{t+1})$} &
  \multicolumn{1}{c|}{0} &
  \multicolumn{1}{c|}{0} &
  \multicolumn{1}{c|}{10} &
  10 \\ \hline
\end{tabular}
\vskip-10pt
\end{table}

The agent is encouraged to track the reference motor position by $r(\mathbf{q}_m, \mathbf{q}^r_m(t))$, pelvis height by $r(q_z, q^r_z(t)+c_z)$, and foot height $r(e_z, e^r_z(t)+c_z)$ at current timestep $t$. 
However, as recorded in Table~\ref{tab:reward}, the reference motion tracking term has a relatively small weight during the multi-goal training because we want the agent to infer diverse maneuvers, such as jumping to different locations, and the jumping-in-place reference motion may be no longer reasonable.

The task completion reward, on the contrary, is designed to dominate others during multi-goal training. 
We first include the $r(q_{x,y}, c_{x,y})$ and $r(q_{\psi,\theta,\phi}, [0,0,c_{\phi}])$ to encourage the agent to reach the desired location and orientation and stay there after it lands in order to accomplish the assigned task $\mathbf{c}$. 
Furthermore, pelvis linear velocity tracking $r(\dot{q}_{x,y}, \dot{q}^d_{x,y})$ and angular rate tracking $r(\dot{q}_{\psi,\theta,\phi}, [0,0,\dot{q}^d_\phi])$ are introduced to shape the sparse task reward, where $\dot{q}^d_{x,y} = c_{x,y}/T_J$ and $\dot{q}^d_{\phi} = c_\phi/T_J$.
Moreover, although the task does not include the pelvis roll and pitch angle $q_{\psi,\theta}$, minimizing them to zero can help to stabilize the robot's pelvis.

We further introduce a smoothing term that is less important than task completion but with a larger weight than the motion tracking term. 
For example, we encourage the robot to produce less ground impact force $F_z$ during its jump by $r(F_z,0)$, to damp the body's oscillation after it lands by motor velocity reward $r(\dot{\mathbf{q}}_m,0)$ and joint acceleration reward $r(\ddot{\mathbf{q}},0)$, and to be more energy efficient by $r(\bm{\tau},0)$. 
Moreover, the importance of having a stationary standing pose is highlighted by having a relatively large weight on torque consumption and motor velocity reward after the robot lands ($t>T_J$). 
This is because the introduction of dynamics randomization in Stage 3 will make the environment noisy and cause oscillation in body pose during standing.
We also introduce an additional component in Stages 2\&3, the change of action reward $r(\mathbf{a}_t, \mathbf{a}_{t+1})$, to further smooth the aggressive maneuver the robot may conduct to jump over a long distance.

\begin{remark} 
Although the switching time ($T_J$) of the reward is fixed and not tuned for jumping to different locations, as we show in our experiments, the robot learns different flight times for different targets (Fig.~\ref{subfig:snapshot_tablex3}).
\end{remark}

\subsection{Episode Design}
Having a careful design of the reward may not be enough since it is challenging to encourage the agent to jump. 
The robot may keep failing to stabilize itself while learning to jump. 
Therefore, the robot may easily adopt very conservative but stable behaviors because it can quickly improve the return in this way. 
For example, the robot may just stand or just jump in place without completing the task, and can still have some suboptimal return. 
To prevent this, we note that a cautious design of the episode can also facilitate the training of dynamic jumping maneuvers. 

In the stages for multi-goal training, the maximum episode length is set to $2500$ timestep which lasts $76$ seconds.
During such an episode, the robot is commanded to jump to a random target after a random time interval of standing, and these random values are uniformly sampled.
The task is sampled from $c_x~\sim~U(-0.5, 1.5)$~m, $c_y~\sim~U(-1.0, 1.0)$~m, $c_z~\sim~U(-0.5, 0.5)$~m, and $c_\phi~\sim~U(-100^\circ, 100^\circ)$, and standing phase distribution is $U(1,15)$ second. 
Such a “jump $\leftrightarrow$ stand" switch is repeated.
Such a design can improve the robustness of the learned policy to different initial states by performing repeat jumps over an episode. 
Moreover, compared to Stage 1 where the agent is asked to jump at $t=0$, starting from Stage 2, there is a high probability the robot will start with a standing skill at each episode.

The episode will be terminated earlier if the robot falls over (pelvis height $q_z < 0.55$ m or the tarsus joints hit the ground) to prevent it from having further rewards. 
We also emphasize the importance of foot height tracking and task completion to the robot by terminating the episode earlier if: (i) the foot height tracking error $|e_z - e^d_z|$ is larger than the bound $E_e$ which is set at $0.32$~m, or (ii) the robot does not arrive at the given target after it lands ($t>T_J$) when $[||q_{x,y}- c_{x,y}||_2, |q_\phi-c_\phi|] > E_t$ where $E_t=[0.35~\text{m}, 35^\circ]$.
Please note that we have a relatively small task completion error bound $E_t$ while we have a large tolerance on the foot height tracking error $E_e$. 
Using such a design, the robot is allowed to deviate from the reference foot trajectory to find a better foot height trajectory for different tasks.
The robot will also have more incentive to complete the task by landing close to the target and having more rewards in a longer episode. 

\begin{remark}
    The larger choice of the foot tracking error tolerance $E_e$ also allows the robot to perform small hops after it lands. The robot is encouraged to stand due to the foot height tracking reward but can dynamically switch to hop, including hopping to different places as long as staying within $E_t$, for better robustness. 
    We do not specifically train or encourage the agent to deviate from the assigned task for robustness. 
\end{remark}

In the first stage of training, a single jump introduces an inductive bias into the policy towards performing jumping behaviors. In later stages of training, by combining the early termination conditions and motion imitation reward, this inductive bias leads the robot to favor jumping to different targets, instead of using other skills such as walking.

\subsection{Dynamics Randomization}~\label{subsec:dynamics_randomization}

In order to succeed during the sim-to-real transfer, we introduce extensive randomization on dynamics parameters of the environment in Stage 3. 
The dynamics properties that are randomized are listed completely in Table~\ref{tab:randomization} in Appendix~\ref{subsec:dynamics_rand_appendix}. 
During training at this stage, at each episode, the value of each dynamics parameter is uniformly sampled from the range listed in Table~\ref{tab:randomization}. 
We consider three sources that cause the sim-to-real gap: (1) modeling errors, (2) sensor noise, and (3) communication delay between the high-level computer running the RL policy and the robot's low-level computer. 

In order to robustify the policy to the modeling errors, we randomize the floor friction, robot's joint damping, link mass and inertia, and the position of the link's Center of Mass (CoM). 
Specifically, to deal with the error of motor dynamics between the simulation and hardware, we have a larger upper bound of the joint damping ($4$ times the default value) to approximate the motor aging issues on the hardware. 
We also randomize the PD gains used in the joint-level PD controllers (since our policy outputs target motor positions). The range is $\pm 30\%$ of the default value. Such a change is able to diversify the motor responses the robot is trained on and enhance the robustness to the change of motor dynamics during hardware deployment. 
Furthermore, specific to Cassie whose leg has leaf springs to connect the passive joints $q_{5,6}$, the parameters of the springs are important because they will have significant displacement during the taking-off and landing phases.
Therefore, we introduce a $20\%$ uncertainty on spring stiffness during training. 
We empirically found that the randomization of the motor dynamics and spring stiffness has a non-trivial effect to succeed during the sim-to-real for the bipedal jumping skills. 

The sensor noise from joint encoders, IMU, and estimation error of the base linear velocity are simulated as a Gaussian noise whose mean is sampled in Table~\ref{tab:randomization} at each episode.

\section{Training Setup}
We now build up our control policy by optimizing reinforcement learning through the multi-stage training pipeline. 
\subsection{Policy Architecture}

\begin{figure}
\centering
\includegraphics[width=\linewidth]{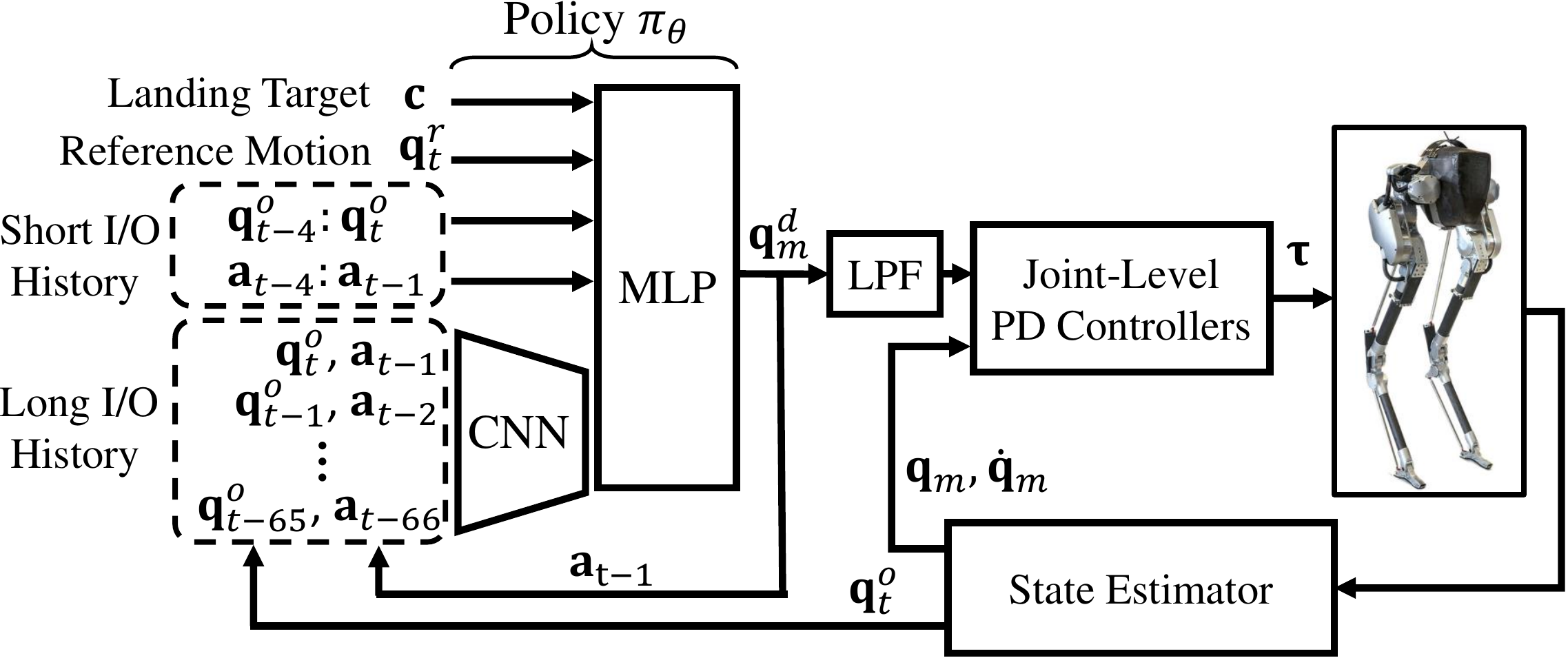}
\caption{The architecture of the goal-conditioned jumping policy $\pi_\theta$. The policy outputs the desired motor positions $\mathbf{q}^d_m$, which are used by joint-level PD controllers to generate the motor torques $\bm{\tau}$ on the robot. The input to the policy includes the goal $\mathbf{c}$, which specifies the landing targets, the reference motion $\mathbf{q}_t^r$, which provides the robot a short preview of the reference trajectory, and a short 4-timestep history of the robot's input (robot's action $\mathbf{a}_{t-1}$) and output (robot's feedback $\mathbf{q}_t^o$).
The policy is also provided with a long-term 2-second I/O history, which is first encoded by a 1D CNN. The policy updates at $33$ Hz while the rest runs at $2$ kHz.}
\label{fig:controller}
\vskip-10pt
\end{figure}

Our policy $\pi_\theta$ is represented by a deep neural network with parameters $\theta$. 
As shown in Fig.~\ref{fig:controller}, it has two components, a base network represented by a multilayer perceptron (MLP), and a long-term history encoder represented by a 1D convolutional neural network (CNN). The policy operates at $33$~Hz. Each action $\mathbf{a}_t$ specifies the target motor positions $\mathbf{q}^d_m$ for the robot.  
The action is first passed through a Low Pass Filter~(LPF)~\cite{peng2020learning,li2021reinforcement,escontrela2022adversarial}, which smooths the motor targets before being applied to joint-level PD controllers and further complements the smoothing rewards. The PD controllers operates at $2$ kHz, to generate motor torques $\bm{\tau}\in \mathbb{R}^{10}$ for driving the movements of the joints.

The input to the policy at timestep $t$ contains four components: the goal $\mathbf{c}$ introduced in Sec.~\ref{subsec:task}, a preview of the reference trajectory $\mathbf{q}^r_t$, a short-term history of previous actions and states (robot's Input/Output), and a long-term I/O history of the last 2-second.
The preview of the reference trajectory $\mathbf{q}^r_t=[q^r_z(t),\mathbf{q}^r_m(t+1), \mathbf{q}^r_m(t+4), \mathbf{q}^r_m(t+7)]$ provided in the robot's observation contains the current reference pelvis height $q^r_z(t)$ and reference motor positions $\mathbf{q}^r_m$ sampled at $1$, $4$, and $7$ future timesteps. 
Providing a segment of the future reference trajectory as input provides the policy with more information such as future joint position, velocity and other higher order terms, which has been used in~\cite{peng2020learning,li2021reinforcement,smith2022legged}.
To close the control loop, we provide the robot direct access to a short-term I/O history of the robot $(\mathbf{q}^o_{t-4:t}, \mathbf{a}_{t-4:t-1})$ in the previous $4$ timesteps (about $0.12$ second).
The I/O history enables the policy to infer the dynamics of the system from past observations. The task $\mathbf{c}$, reference motion $\mathbf{q}^r_t$, and the short-term I/O history at current timestep $t$ are directly passed as inputs to the base MLP.

For sim-to-real transfer, a short-term history may not be enough to provide adequate information to control dynamic maneuvers on a high-dimensional system. 
For example, during a jump, the landing event is affected by the angular momentum gained before the take-off, and the interval between these two events can be much longer than the $0.12$ second.
Therefore, we include an additional input in the form of a long-term I/O history of the past $2$ seconds, which contains $66$ timesteps of past observations and actions measurements $(\mathbf{q}^o_{t-65:t}, \mathbf{a}_{t-66:t-1})$. 
The timespan of this long I/O history is designed to cover the duration of a jump to help the policy implicitly infer the robot's dynamics, traveled trajectory, and contacts.
To encode this long sequence of observations, we use a 1D CNN to compress it into a latent representation before providing it as an input to the base MLP. 
As we will see in Fig.~\ref{fig:learning_logs}, \emph{both} the long-term and short-term history are needed for better learning performance.

In this work, the CNN encoder consists of 2 hidden layers whose [kernel size, filter size, stride size] are $[6, 32, 3]$ and $[4, 16, 2]$ with \textit{relu} activation and no padding, respectively.
The result of the CNN is flattened and concatenated into the inputs of the base MLP. 
The MLP has two hidden layers with $512$ \textit{tanh} units, followed by an output layer representing the mean of a Gaussian action distribution with a fixed standard deviation of 0.1$I$.

\subsection{Training Details} Empirically, we found that simultaneously performing both turning and jumping to different elevations is very difficult for Cassie, which does not have a torso. 
Due to this hardware limitation, we choose to train two separate goal-conditioned policies: a \textit{flat-ground policy} that is specialized for jumping without elevation changes, \textit{i.e.}, $c_z=0$, and a \textit{discrete-terrain policy} that is trained to jump onto platforms with different elevations without turning ($c_\phi=0$).

Proximal Policy Optimization~(PPO)~\cite{schulman2017proximal} is used to train all policies $\pi_\theta$ in simulation, with a value function represented by a 2-layered MLP, which has an access to the ground truth observations. 
Due to the differences in the complexity of the different training stages, the three stages are trained with 6k, 12k, and 20k iterations, respectively. Each iteration collects a batch of 65536 samples.

\section{Simulation Validation}
Having introduced our methodology for training goal-conditioned jumping policies,
we will next validate the proposed method in simulation (MuJoCo). 
In this section, we aim to address two questions: (1) what are the advantages of the proposed policy architecture compared to models used in prior work, (2) whether training with multiple tasks can further improve the robustness of the policy over single-goal training, by allowing the robot to utilize more diverse maneuvers to recover from unstable states or unknown perturbations.

\subsection{Baselines}

\begin{figure}[t]
\centering
\includegraphics[width=\linewidth]{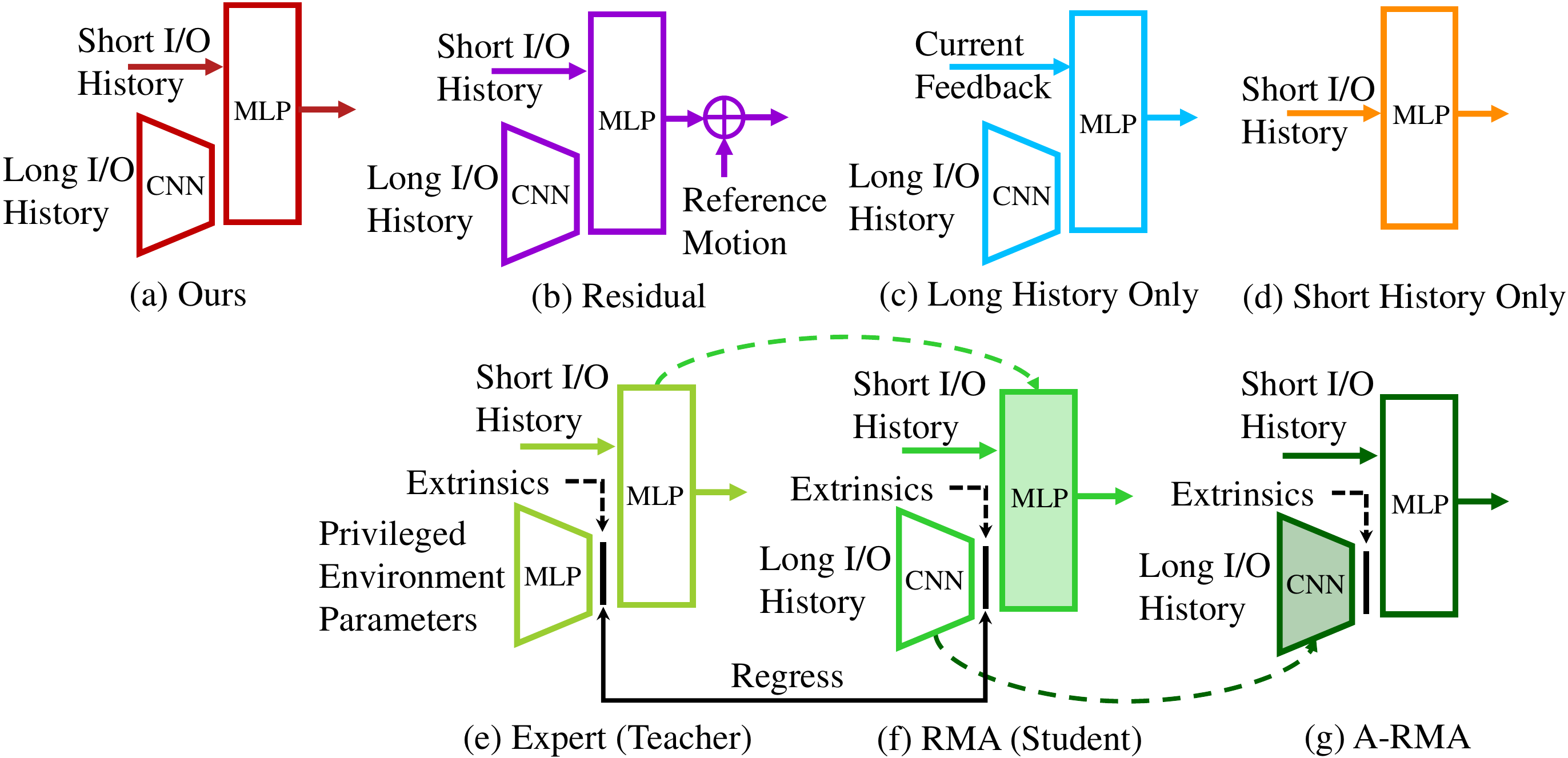}
\caption{Illustration of the baseline policy structures used to train the policy for bipedal jumping. (a) Ours: proposed structure as discussed in detail in Fig.~\ref{fig:controller}. (b) Residual policy that has the same input structure as our method but outputs a residual term adding to the reference motor position~\cite{lee2020learning,xie2020learning}. (c) Long History Only policy that only has the access to a long-term I/O history (we still provide robot immediate feedback to the base, as suggested by~\citet{peng2018sim}).
(d) Short History Only policy that is only provided with a short-term I/O history~\cite{li2021reinforcement}. We also compare with the RMA~\cite{kumar2021rma}/Teacher-Student~\cite{lee2020learning} training strategy where an (e) expert policy with access to privileged environment information (Table~\ref{tab:randomization}) is first trained by RL and is later utilized to train (f) RMA (student) policy by supervised learning. The RMA can be further finetuned using (g) A-RMA~\cite{kumar2022adapting} by RL. While the short I/O history is not included in the original RMA~\cite{kumar2021rma} or TS~\cite{lee2020learning}, it is included in this benchmark to have a fair comparison. The blocks are shaded if their parameters are not updated. The dash lines indicate that parameters are copied.}
\label{fig:benchmark}
\vskip-10pt
\end{figure}

To answer the first question, we benchmark our proposed policy architecture with several baselines illustrated in Fig.~\ref{fig:benchmark}. 
All policies are trained with multiple goals, \textit{i.e.}, jumping to different landing locations and turning directions with no change of elevation, using the training schematic shown in Fig.~\ref{fig:pipeline}, and are trained with $3$ different random seeds. 
The details of baseline models are described in Appendix~\ref{subsec:baseline_appendix}. 

To address the second question, we obtained two single-goal policies using the proposed policy structure, as detailed below:
\begin{itemize}[leftmargin=9pt]
    \small
    \item \textbf{Single Goal}: a policy that is trained on a single jumping-in-place task and extensive dynamics randomization as listed in Table~\ref{tab:randomization}.
    \item \textbf{Single Goal w/ Perturbation}: a policy similar to the single-goal policy but is also trained with a randomized perturbation wrench (6 DoF) applied on the robot pelvis. The external forces and torques are sampled uniformly from $[-20\text{N}, -5\text{Nm}]$ to $[20\text{N}, 5\text{Nm}]$ and are applied on the robot's pelvis for a random time interval ranging from $[0.1, 2.0]$ second. 
\end{itemize}

We compare these baselines based on two metrics: (1) learning performance in Sec.~\ref{subsec:learning_performance} and (2) the ability to generalize to dynamics parameters that lie outside of the training distributions in Sec.~\ref{subsec:single_task}. 
These two metrics are important for the sim-to-real transfer because the first one shows how well the policy can perform during training
and the second evaluates robustness to changes in the environment, which are not considered during training, as can be the case during sim-to-real transfer.

\subsection{Policy Structure Choice}~\label{subsec:learning_performance}

\begin{figure}[t]
\centering
\includegraphics[width=0.78\linewidth]{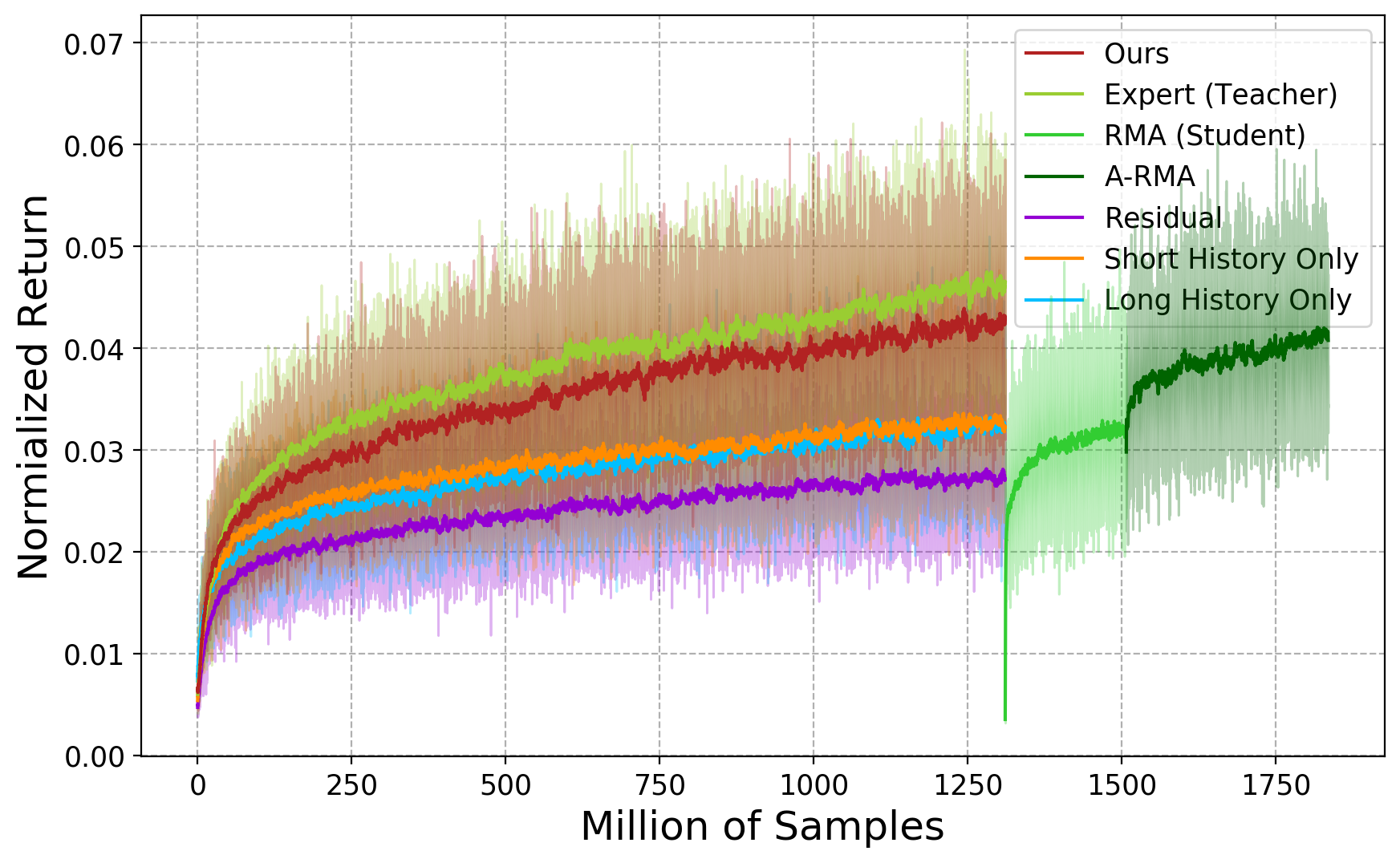}
\caption{Benchmark of learning curves trained by different policy structures in Stage 3 (multi-goal training with dynamics randomization). The curves are the average normalized returns trained with $3$ random seeds while the colored areas enclose the min and max values obtained among different seeds. The normalized return is calculated by the return divided by the max episode length and in the range of $[0,1]$. Our method shows similar performance as the expert policy which is used to supervise RMAs and has access to the privileged environment parameters. The A-RMA shows the second-best performance but it requires significantly more samples compared to the proposed methods, followed by RMA. The policies with short history only or long history only show a similar learning performance but are a bit worse than RMA in terms of returns. The residual policy shows the worst performance because the reference motion added to the policy's action prevents the agent from exploring more diverse maneuvers.}
\label{fig:learning_logs}
\vskip-10pt
\end{figure}

The learning curves from Stage 3 (multi-goal learning with dynamics randomization) using our policy structure and baselines are presented in Fig.~\ref{fig:learning_logs}. The learning curves at early training stages (learning a single task in Stage 1 and multiple tasks in Stage 2) are available in Fig.~\ref{fig:learning_logs_appendix} in Appendix~\ref{subsec:curve_stage12_appendix}.
The same hyperparameters and reward functions are used for every training stage.


According to Fig.~\ref{fig:learning_logs} (and Fig.~\ref{fig:learning_logs_appendix}), the residual structure drawn as the purple curve shows the worst learning performance over all the training stages. 
The reason is the reference motion we provided is a dynamically-infeasible animation, which may cause the robot to spend more effort learning to correct these default motions, and prevents it from exploring more diverse trajectories and inferring the motion that is outside of the range of the reference motion.

\begin{figure*}[t]
\centering
\begin{subfigure}{0.4425\linewidth}
  \centering
  \includegraphics[width=\linewidth]{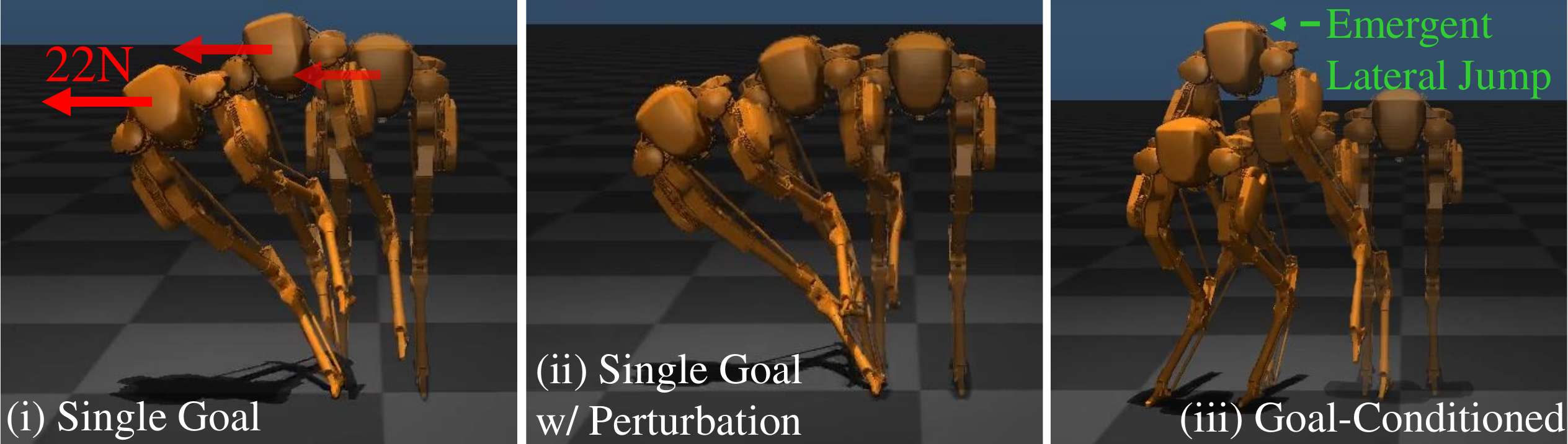}
  \caption{With Consistent Unknown Lateral Perturbation Force}
  \label{subfig:bm_single_sim_perturb}
\end{subfigure}
\begin{subfigure}{0.5425\linewidth}
  \centering
  \includegraphics[width=\linewidth]{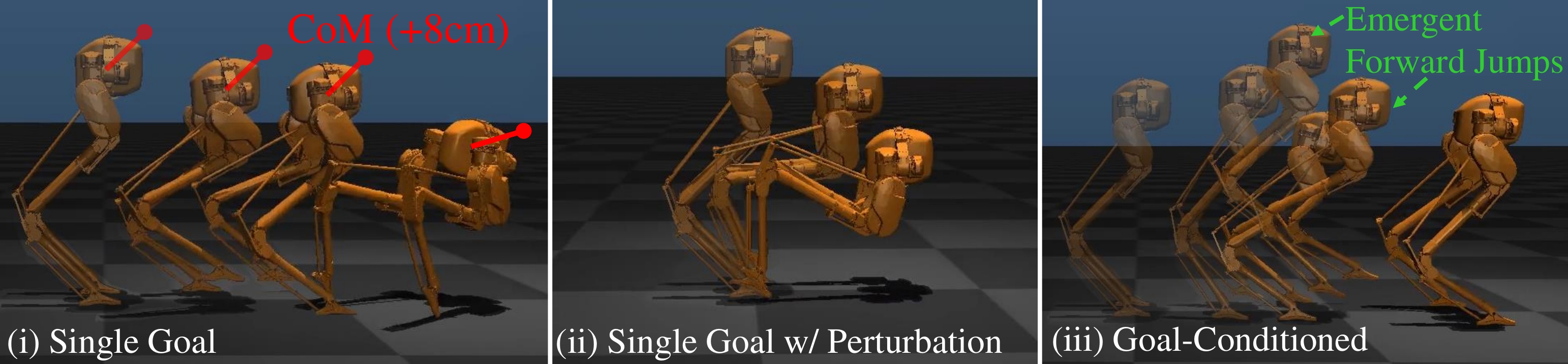}
  \caption{With Errors in Center of Mass Positions of All Links}
  \label{subfig:bm_single_sim_ipos}
\end{subfigure}
\caption{Robustness comparison among three policies which are: (i) trained with a single task (jumping in place) with dynamics randomization, (ii) trained with a single task with dynamics randomization and random perturbation, and (iii) trained with multiple tasks with dynamics randomization but without random perturbation (proposed). The testing scenarios are outside the training setting for all three policies. The single-goal policies fail to stabilize the robot, even the one trained with extensive perturbations. The goal-conditioned policy which is trained with diverse jumping tasks but without perturbation succeeds to stabilize the robot by exploiting the learned skills. The goal-conditioned policy is able to deviate from the commands (jumping in place) and utilize a lateral jump to stay robust to the lateral external force and two forward jumps to adapt to the forward CoM offset.}
\label{fig:bm_single_sim}
\vskip-10pt
\end{figure*}

The baselines using short history only (orange curve) and long history only (blue curve) show a similar learning performance.
But if we combine these two by providing the policy with a long history encoder \emph{and} direct access to short history, which results in our method, the learning performance can be enhanced to a large extent, as drawn as the red curves in Fig.~\ref{fig:learning_logs}. 
This showcases that, providing the policy with a long history is not enough because the robot may need immediate feedback which may be hidden from the long-history encoder. 
Providing the policy with direct access to the short history can address it and the agent can learn to utilize both information.

\begin{remark}
We note that there is other work using RNNs with LSTM~\cite{peng2018sim,siekmann2020learning,shao2021learning} or TCN~\cite{lee2020learning} to encode the long-term I/O history. We hypothesize that providing the policy direct access to the \emph{short history} is not limited to the 1D CNN encoder but also to other neural network structures that capture temporal information such as TCN, LSTM, GRU, and Transformer. We choose 1D CNN in this work because it is easier to train. 
\end{remark}

The comparison between our method and RMA/Teacher-Student (TS) policies (green curves) is also interesting. 
During the training of the goal-conditioned policy with dynamics randomization, our method only shows a little degradation compared to the expert policy. 
This actually showcases the advantages of our method because it can be zero-shot transferred to the real world while the expert policy that requires privileged information cannot.
After training the expert policy, RMA and A-RMA have trained with 3k iterations and 5k iterations respectively, as shown in Fig.~\ref{fig:learning_logs}. 
We found that RMA has a large degradation compared to the expert policy due to the regression loss, and A-RMA is necessary to finetune the base policy in order to further improve the return.
RMA shows a better return than the policy with only short history or only long history, which is aligned with the finding from previous work~\cite{kumar2021rma,kumar2022adapting}.
However, even after A-RMA converged, the return is a bit worse than our method, while RMA and A-RMA require additional training and significantly more samples. 

\begin{remark}
    The original implementation of RMA~\cite{kumar2021rma} or TS~\cite{lee2020learning} only provides the robot's very last I/O pair~\cite{kumar2021rma} or last state feedback~\cite{lee2020learning} besides the long-term history encoder. In the implementation of RMA/TS in Fig.~\ref{fig:benchmark}, we added the short-term I/O history, which can improve the learning performance in order to have a fair comparison. 
    Furthermore, the long-term I/O history encoder used in RMA and A-RMA is the same as the one used in the proposed method, which shows a better learning performance than the original encoder proposed in~\cite{kumar2021rma,kumar2022adapting} as shown in Fig.~\ref{subfig:rma_encoders_logs_appendix}.  
\end{remark}

\textit{Summary of the Result}:
By the ablation study above, we can summarize three factors that can improve the learning performance in our case for dynamic locomotion control: (1) using desired motion positions as the action space (in contrast to the residual), (2) providing the policy with direct access to the short-term I/O history in addition to a long-term robot's I/O history, and (3) training the policy in an end-to-end way instead of separating the training process into teacher and student. This combination leads to our proposed method.

\subsection{Advantages of the Verstiale Policy}~\label{subsec:single_task}

\begin{figure*}
\centering
\begin{subfigure}{0.495\linewidth}
  \centering
  \includegraphics[width=\linewidth]{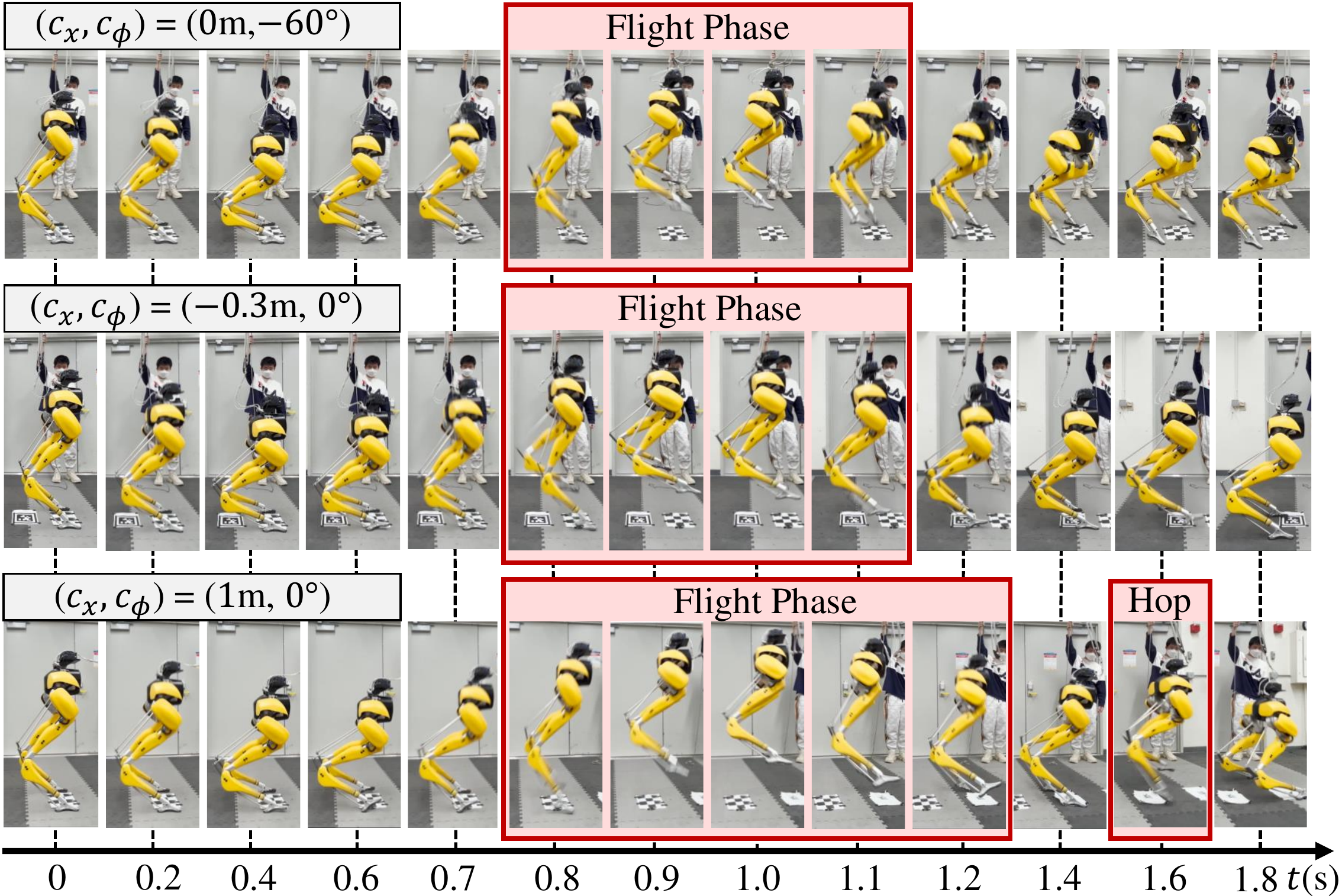}
  \caption{Different Jumps using the Flat-ground Policy}
  \label{subfig:snapshot_flatx3}
\end{subfigure}
\begin{subfigure}{0.495\linewidth}
  \centering
  \includegraphics[width=\linewidth]{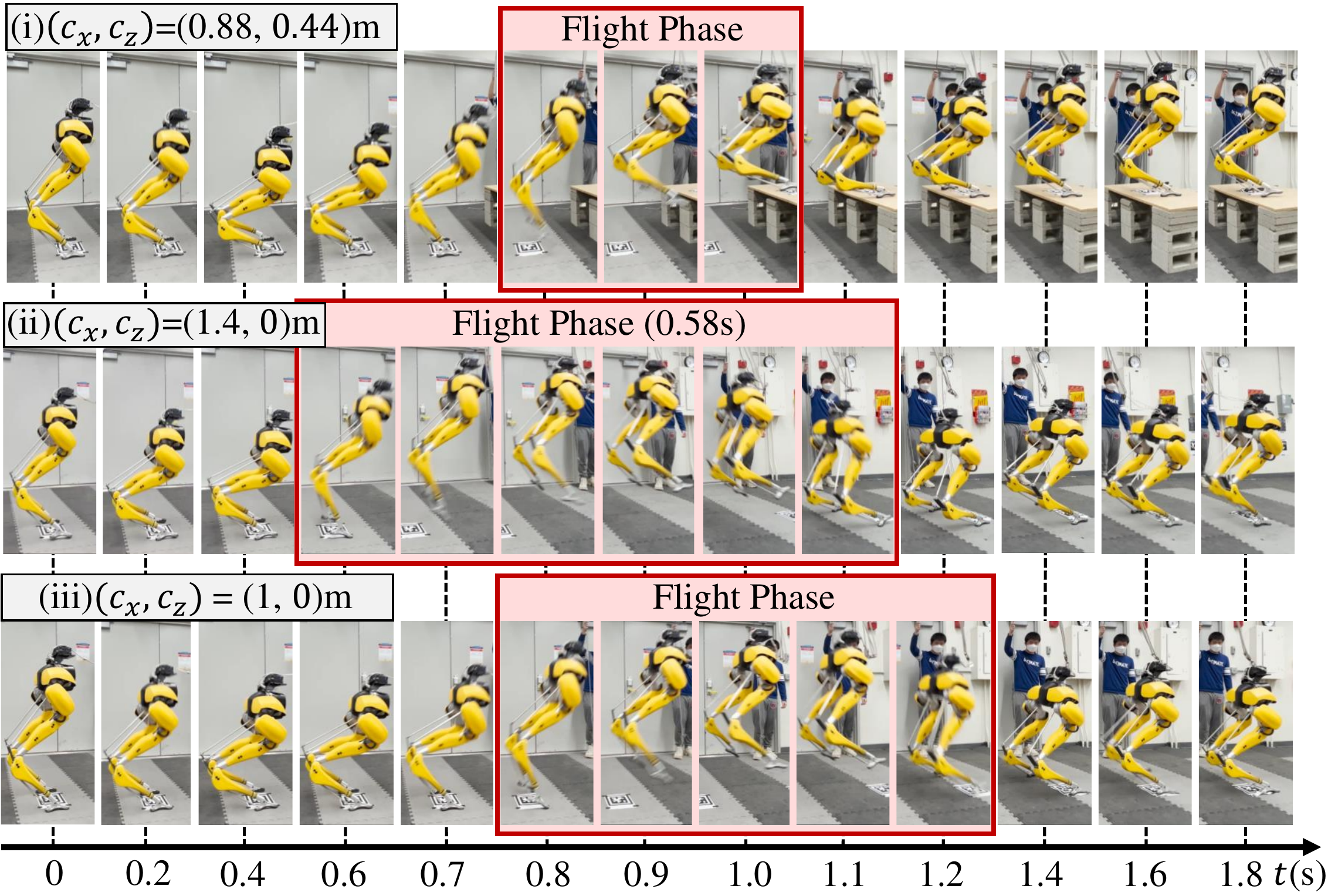}
  \caption{Different Jumps using the Discrete-terrain Policy}
  \label{subfig:snapshot_tablex3}
\end{subfigure}
\caption{Snapshots of Cassie performing different jumps using the proposed goal-conditioned policies. The snapshots are aligned with timestamps. The tags in the figures indicate the given landing targets. (a) Using a single policy that is specialized on flat ground, the robot is able to (i) jump in place while turning to $-60^\circ$, (ii) jump $0.3$~m backward, and (iii) jump $1$~m forward, respectively. During the $1$~m jump, the robot utilizes a forward hop to reach the goal after it lands on $0.5$~m after the first jump. (b) The robot utilizes a single discrete-terrain policy to jump to different locations and elevations. The single policy can change the contact plan for different tasks. For example, the flight phase of the $1.4$~m forward jump (ii) is the longest while being the shortest when the robot jumps onto the $0.44$~m high elevation (i). The robot can land at the target (tag) with insignificant errors among all of these jumps without global position feedback.}
\label{fig:snapshot_timeline}
\vskip-10pt
\end{figure*}

In order to validate the advantages brought by multi-goal training, we further compare our goal-conditioned policy with the single-goal policies. 
These two single-goal policies are trained with the same amount of samples with dynamics randomization as the proposed one, whose learning curves are recorded in Fig.~\ref{subfig:single_logs_appendix}.
During the test in simulation, we command the robot to perform an in-place jump in an environment that the robot has not been trained on. 
As presented in Fig.~\ref{fig:bm_single_sim}, we conducted two tests where (1) a consistent lateral perturbation force is applied on the robot pelvis, and (2) the CoM of \textit{all} links are set to be $+8$ cm off from the default position in all dimensions, while other dynamics parameters are set to the default values.  

During these two tests, both of the single-goal policies fail to control the robot, while the goal-conditioned policy succeeded to stabilize the robot and perform a jump. 
Specifically, the policies trained with a single goal directly fail during standing, even in the case where one is trained with extensive external perturbations that “force" the robot to explore more maneuvers by perturbing it from a nominal jump.  
On the contrary, the policy trained with multiple goals, such as jumping forwards and lateral, without perturbations during training, is able to generalize the learned tasks, exploit them to stabilize the robot, and pick the best jumping maneuver even if it is not commanded. 
For example, while being commanded to jump in place, the goal-conditioned policy utilizes a lateral jump that it learned to stabilize the robot with the presence of lateral force (Fig.~\ref{subfig:bm_single_sim_perturb}(iii)) and two emergent forward jumps to adapt to the CoM errors in the forward direction (Fig.~\ref{subfig:bm_single_sim_ipos}(iii)). 
Such a benchmark highlights the advantages of learning with multiple tasks which makes the policy more robust. 

Having conducted an extensive ablation study in simulation, we show that the proposed policy structure and multi-goal training significantly improve the robustness of the policy over other policy structures or single-goal policies.

\section{Experiments}

We now deploy the goal-conditioned policies obtained in simulation, the \textit{flat-ground policy} that is trained on different goals to jump to various locations and turning directions, and the \textit{discrete-terrain policy} that is specialized in jumping to variable locations and elevations, on the hardware of Cassie. 
As shown in Fig.~1, both policies can successfully control the robot in the real world, without finetuning. 

Besides the ability to succeed in sim-to-real, in this section, we aim to validate two hypotheses: 1) whether the policy trained in simulation can complete the same task in the real world, and 2) whether the goal-conditioned policy is still able to exploit the learned tasks to stabilize the robot after being transferred to the real world. 
The experiments can be best seen in the accompanying video (\url{https://youtu.be/aAPSZ2QFB-E}). 
Please note that in all of the experiments, the robot does not have global position feedback, \textit{i.e.}, once it starts to move, it does not know the distance to the landing target nor the distance to the ground.

\subsection{Task Completion in the Real World}
We first test the flat-ground policy on the robot in three distinct tasks: jumping in place while turning to negative $60$ degrees, jumping $0.3$~m backward, and jumping forward to land at a $1$~m ahead target. 
As recorded in Fig.~\ref{subfig:snapshot_flatx3}, controlled by this goal-conditioned policy, the robot is able to complete all three tasks. 
For example, during the turning task, the robot rotates to $-55^\circ$ while jumping in the air, and lands at the same place where it took off (marked by a tag on the ground in Fig.~\ref{subfig:snapshot_flatx3}(i)). 
During the backward jump, different from the previous task, the robot leans backward before taking off ($0.6$-$0.7$ sec in Fig.~\ref{subfig:snapshot_flatx3}(ii)), and lands accurately at the target tag on the ground. 
To jump $1$~m forward, the robot adopts a different maneuver where it leans forward before the flight phase and pushes itself off the ground with a larger strength, which results in a longer flight phase and travel distance than the previous two tasks. 
We also observe that the robot first lands at a $0.5$~m landmark, but quickly conducts a forward hop when legs touch the ground ($1.6$~sec in Fig.~\ref{subfig:snapshot_flatx3}(iii)), and lands at the $1$~m target tag in the last. 
We note that such a consecutive jumping maneuver does not happen during the same task in the simulation with the robot's nominal dynamics model.
Such an experiment highlights two favorable features of the proposed policy: it can (1) adapt to different system dynamics (from sim to real) and (2) deviate from the reference motion and utilize multiple contacts to complete the given task (jumping to the target).   

\begin{figure*}
\centering
\includegraphics[width=0.8\linewidth]{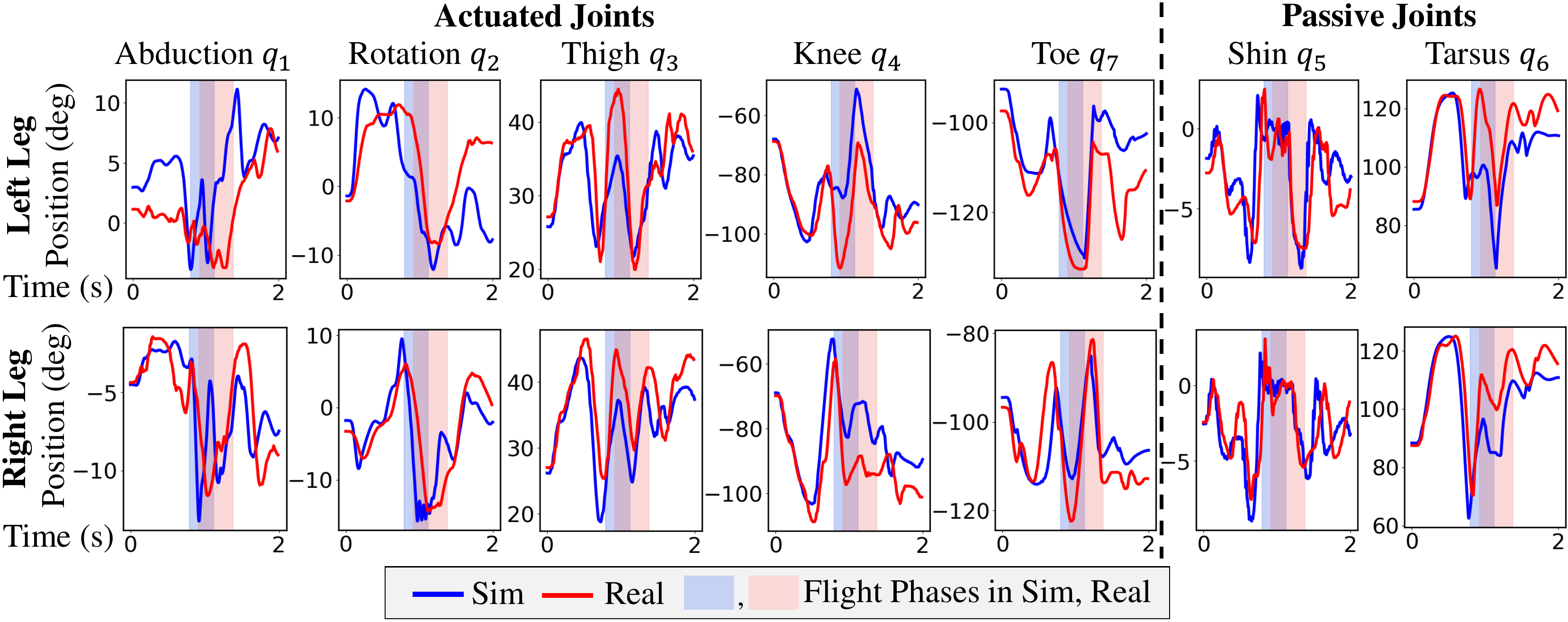}
\caption{The profiles of the robot's joint positions when it is commanded to jump and turn $-60^\circ$ in place in simulation and the real world. We observe a large deviation of the joint profiles between sim and real, \textit{e.g.}. the tarsus joints which are passive and driven by leaf springs show a significant difference during the sim-to-real transfer. Moreover, the flight phase in the real world is delayed compared with the one in the sim. Such errors highlight a big sim-to-real gap but our policy is robust to this and succeeds in controlling the robot to the given target.}
\label{fig:logs}
\vskip-10pt
\end{figure*}

We then validate the discrete-terrain policy with three tasks: Jumping $1$~m ahead, $1.4$~m ahead, and to a target that is $0.88$~m ahead and $0.44$~m above the ground, as presented in Fig.~\ref{subfig:snapshot_tablex3}. 
It shows the capacity to control the robot to jump over a distance/elevation to land on the given target accurately. 
We notice that the policy is able to adjust the robot's maneuvers/contact plan to jump over different distances/elevations. 
For example, compared to the $1$~m jump (Fig.~\ref{subfig:snapshot_tablex3}(iii)), the robot takes off earlier while landing later during the $1.4$~m jump (Fig.~\ref{subfig:snapshot_tablex3} (ii)). 
Such a change is reasonable as the robot needs a longer flight phase/larger take-off velocity in order to land at a farther location.
Furthermore, when the robot is commanded to jump onto a $0.44$~m table, the policy jumps more vertically ($0.8$~sec in Fig.~\ref{subfig:snapshot_tablex3}(i)) compared to the $1.4$~m jump ($0.6$~sec in Fig.~\ref{subfig:snapshot_tablex3}(ii)) at the beginning of the flight phase while lifting the robot legs much higher in order to jump higher. 
Note that the robot makes contact with the platform much earlier than the other jumps on the ground, but this single policy is still able to stabilize the robot with different landing events.
Furthermore, all these three experiments show that the robot controlled by the proposed policy can land accurately on the given target (land on the tags in Fig.~\ref{subfig:snapshot_tablex3}), which is challenging. 
Because the robot's motion is ballistic in the flight phase and a small error during taking-off may result in a large deviation from the landing target. 
In these experiments, the proposed policy is able to adapt to the dynamics of the robot hardware, adjust the robot's pose during the take-off, and accelerate to a velocity that can land the robot on the target. 

\begin{remark}
    During the long jump (like Fig.~\ref{subfig:snapshot_tablex3}(ii)), the robot leans its body forward at a large angle when it is pushing off from the ground and swings the legs forward during descent, and rotates its body forward w.r.t. contact points after it lands. Such a maneuver is very close to what we observed when a human athlete performs a standing jump~\cite[Fig. 1A]{moresi2011assessment}. Similar to humans, our robot's long jump skill is also learned during training, which is very different from the jumping-in-place reference motion we provided.  
\end{remark}

\subsection{Sim-to-Real Gap}
In order to further understand the difficulty to succeed in robot jumping experiments in Fig.~\ref{fig:snapshot_timeline}, we take a close look at the sim-to-real gap.
We record the robot's joint position profiles during a jump in the simulation with the robot's nominal dynamics parameters and on the robot's hardware.
The profiles for a turning task ($-60^\circ$, Fig.~\ref{subfig:snapshot_flatx3}(i)) using the flat-ground policy is presented in Fig.~\ref{fig:logs}. 
According to the recorded profiles, the robot's actual joint position has a large deviation between the simulation (blue curves) and the real world (red curves). 
For example, the maximum error on the tarsus joint position $q_6$ between sim and real is over $0.35$~rad, which largely affects the robot's dynamics considering this joint is not actuated and is driven by a leaf spring whose nominal stiffness is $1250$ Nm/rad. 
A similar deviation is observed in other joints, such as rotation joints $q_2$, thigh joints $q_3$ and knee joints $q_4$, which play a critical role during a jump and turning, and in other experiments using the discrete-terrain policy as recorded in Fig.~\ref{fig:table_logs_appendix}.
Such a discrepancy highlights the huge gap between the simulation and the real world but also showcases that, despite such a large gap, our methodology introduced in Sec.~\ref{sec:methodology} is able to stay robust and succeed in controlling the robot to accomplish the task.

\begin{figure*}
\centering
\begin{subfigure}{\linewidth}
  \centering
  \includegraphics[width=\linewidth]{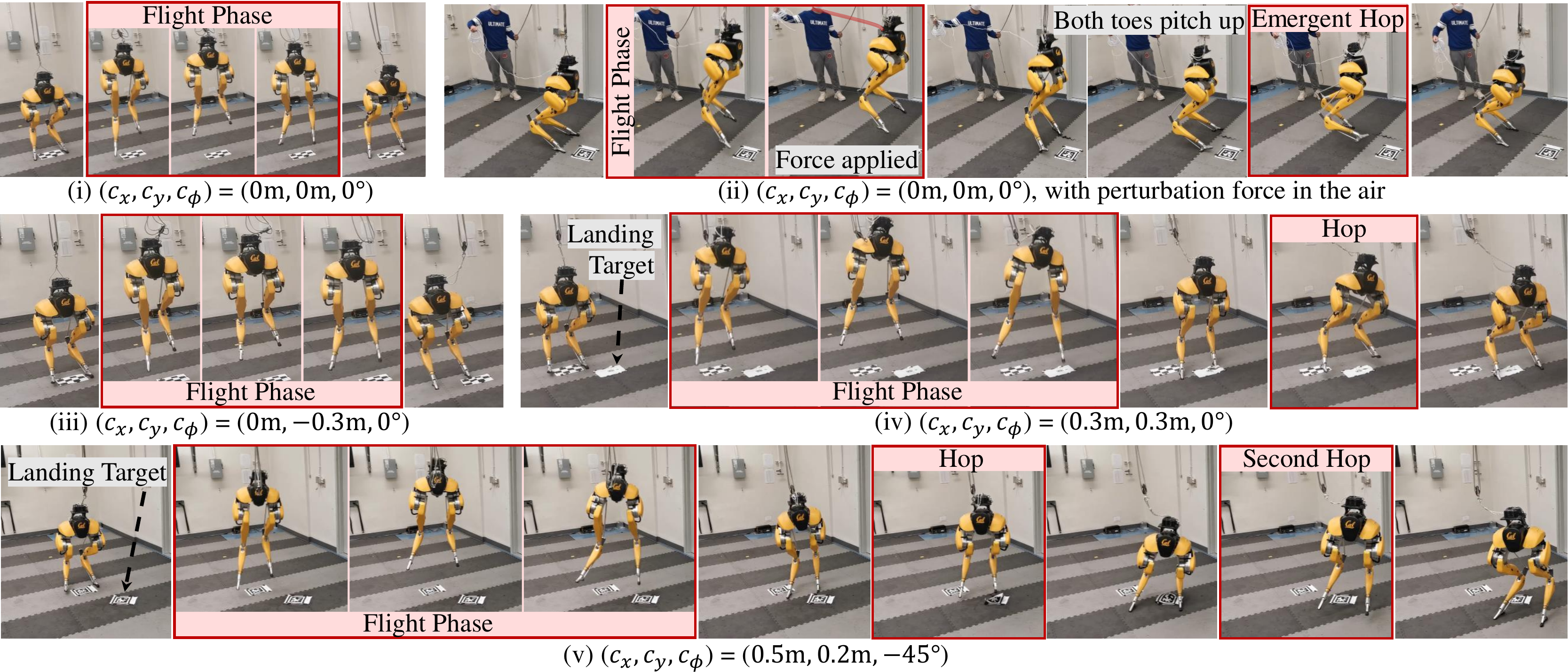}
  \caption{Different Jumps using the Flat-ground Policy}
  \label{subfig:snapshot_flatground_rest}
\end{subfigure}
\begin{subfigure}{\linewidth}
  \centering
  \includegraphics[width=\linewidth]{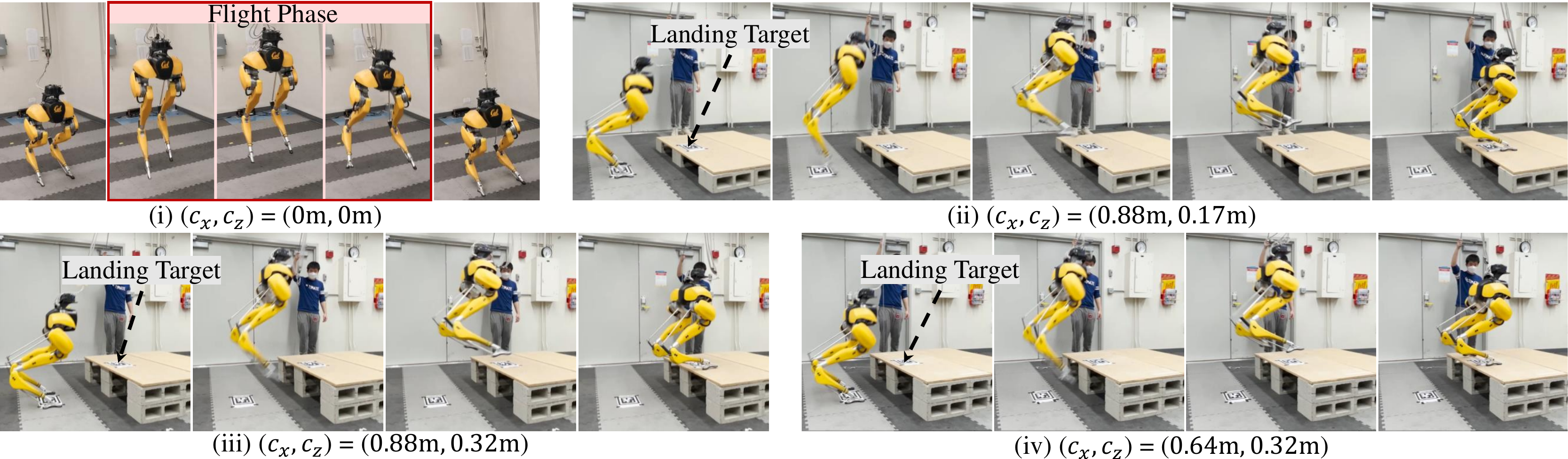}
  \caption{Different Jumps using the Discrete-terrain Policy}
  \label{subfig:snapshot_table_rest}
\end{subfigure}
\caption{Snapshots of various dynamic jumps performed by Cassie using the proposed policies. (a) The robot is able to perform a large repertoire of multi-axes jumps on flat ground. It shows the ability to stabilize the robot from a backward external perturbation (ii) by deviating from the commanded in-place jump and exploiting the maneuvers learned from backward jumping tasks. The robot also leverages emergent hops after landing to stabilize it from a huge impact force, while being commanded to stand, like (iv) and (v). (b) Using a single discrete-terrain policy, the robot can not only jump in place (i) but also jump to different locations with different elevations (ii) (iii) (iv).}
\label{fig:snapshot_rest}
\end{figure*}

\subsection{Diverse and Robust Maneuvers by the Goal-Conditioned Policy}

In order to push the limits of the proposed control policies, we further conduct more dynamic jumping experiments as presented in Fig.~\ref{fig:snapshot_rest} and Fig.~\ref{fig:snapshot_flat_rest_appendix}.
As shown in Fig.~\ref{subfig:snapshot_flatground_rest}, using the single flat-ground policy, the robot performs a large repertoire of dynamic jumping maneuvers, such as jumping in place (Fig.~\ref{subfig:snapshot_flatground_rest}(i)), jumping to lateral (Fig.~\ref{subfig:snapshot_flatground_rest}(iii)), and multi-axes jumps such as blending lateral and forward jumps (Fig.~\ref{subfig:snapshot_flatground_rest}(iv)) and forward, lateral and turning (Fig.~\ref{subfig:snapshot_flatground_rest}(v)). 
In these multi-axes jumps, the robot demonstrates more complex maneuvers. For example, the robot leans in the lateral direction while jumping forward and turning to land on the target that is $0.5$~m ahead, $0.2$~m to robot's left, and turned $-45^\circ$, as shown in Fig.~\ref{subfig:snapshot_flatground_rest}(v).
During some challenging tasks, the robot is aware to utilize small hops to adjust its body pose after it lands with unstable states, such as demonstrated in Fig.~\ref{subfig:snapshot_flatground_rest}(iv)(v).

Moreover, in order to test the robustness of the policy, we applied a backward perturbation force on the robot's pelvis at its apex jumping height, as shown in Fig.~\ref{subfig:snapshot_flatground_rest}(ii). 
Due to such a perturbation, the robot leans backward during descending, and both of its toes pitch up after it lands, which makes the robot underactuated w.r.t contact points. 
However, the robot quickly exerts a backward hop, which is learned during the multi-goal training, after it lands.
By this hop, the robot can adjust its body pose during the flight phase and then land stably afterward.
The goal we gave to the robot in this test is to jump in place and it is interesting to see that the robot deviates from it in order to recover from falling over. 

\begin{remark}
    The robot, controlled by the proposed jumping policy, shows the ability to not rely on the pre-defined contact plan and can break the contact after it lands and make contact again when it needs to utilize impacts to stabilize itself. 
    Such a capability is similar to contact implicit trajectory optimization~\cite{posa2014direct,chatzinikolaidis2020contact,zhu2021contact,drnach2021robust,landry2022bilevel}. While such optimization schemes still need to be computed offline for legged robots, our work achieves this online.
\end{remark}

In the additional testing of the discrete-terrain policy demonstrated in Fig.~\ref{subfig:snapshot_table_rest}, the robot shows the ability to accurately land on different given targets. 
While the changes in the commanded distance and elevation are relatively small, the policy still demonstrates the ability to adjust the robot's take-off maneuvers in order to jump to the given targets.

\section{Discussion of Design Choices for RL-based Legged Locomotion Control}
In this section, we discuss the lessons learned through the development of jumping controllers for bipedal robots using RL. We hope this can provide useful insights for future endeavors on applying RL for legged locomotion.

\textbf{Short-term history complements the long-term history}:
Providing a long-term history of the robot's input (policy's action) and/or output (measurement feedback) has been used in many prior efforts in RL-based robotic controls~\cite{peng2018sim,lee2020learning,kumar2021rma,shao2021learning,siekmann2020learning}. 
While these prior systems show the advantages of using a long-term history over only the current state feedback~\cite{peng2018sim,siekmann2020learning}, the advantages over a short history~\cite{peng2020learning,li2021reinforcement,miki2022learning} were not investigated. 
In this work, we demonstrate that incorporating \textit{both} short-term history and long-term history can be beneficial.
The ablation study in Fig.~\ref{fig:learning_logs} shows that providing only the long-term I/O history (Fig.~\ref{fig:benchmark}c) may not be sufficient, even when combined with observations of the robot's current state besides the history encoder, which is analogous to~\cite{peng2018sim,lee2020learning,kumar2021rma}.
The learning performance of such a method shows no significant difference between the MLP policy with only short-term history (Fig.~\ref{fig:benchmark}d), as shown in Fig.~\ref{fig:learning_logs}. 
Our architecture (Fig.~\ref{fig:benchmark}a) exhibits better learning performance because it has direct access to a short-term history while also having a long-term history encoder.  
During real-time control, the robot's recent feedback and policy outputs (I/O) could be more important than the observations that are further back in time. 
Although it is also part of the long-term history, such recent information can be obfuscated and hard to extract from the compressed latent representation from the long-term history encoder. 
The short-term history provides the model with direct access to the most recent observations.
Therefore, one of the reasons the proposed method shows the best learning performance in Fig.~\ref{fig:learning_logs} is not the usage of long-term history, but the combination of short-term and long-term history. In addition to jumping, this design decision may also benefit other locomotion skills.

\textbf{Encode environment parameters or robot's I/O history?} Although the proposed architecture (Fig.~\ref{fig:benchmark}a), with the exception of the introduction of short I/O history, may resemble the architecture of RMA~\cite{kumar2021rma} or Teacher/Student~(TS) framework~\cite{lee2020learning} which also has a long-term history encoder (Fig.~\ref{fig:benchmark}e,f), the objective of the temporal encoder in this work is different from those methods~\cite{kumar2021rma,lee2020learning}. 
The long I/O history encoder in RMA or TS is to estimate the \textit{human-selected} environment parameters (\textit{e.g.}, floor friction and robot's model parameters) by matching the predicted extrinsics from the teacher policy. 
The proposed method, in contrast, jointly trains the long I/O history encoder with the base policy and learns to directly utilize the robot's past I/O trajectories for control. 
The advantage is, the robot's I/O history implicitly contains more information besides environment parameters, such as impact events and contact wrenches. 
In this way, the robot has more freedom to extract the information from the long I/O history without being restricted to estimating the pre-selected environment parameters. 
This is the reason that the proposed method shows improvement over RMA/TS, which separates training into different teacher and student stages, and also requires an additional finetuning stage by A-RMA, which requires more training time and data, as shown in Fig.~\ref{fig:learning_logs}. 
We would like to also note that, RMA/TS methods potentially have benefits when combined with external sensors like vision~\cite{agarwal2022legged,miki2022learning}, which the proposed one may not have.

\textbf{Robustness comes from versatility}: 
We have observed that some of the prior RL-based locomotion controllers on periodic walking skills show highly robust behaviors during real-world deployment, including being robust to external perturbations~\cite{li2021reinforcement,margolis2022walk} or change of terrains~\cite{kumar2021rma,miki2022learning,lee2020learning}. For aperiodic dynamic jumping skills studied in this work, the RL-based policy also demonstrates significant robustness such as Fig.~\ref{subfig:snapshot_flatground_rest}(ii). 
This phenomenon raises an interesting question: where does the robustness come from and how can we improve robustness when we are using RL for legged locomotion control? 
While there has been little prior work that studies this source of robustness, in this work, we conduct an ablation study in Sec.~\ref{subsec:single_task}. 
Fig.~\ref{fig:bm_single_sim} clearly shows that one source of this robustness stems from multi-goal training: the versatile RL-based policy learned from different jumping tasks is able to generalize the learned task to recover from the unexpected perturbation (Fig.~\ref{subfig:bm_single_sim_perturb}(iii)) or deviation from the nominal trajectory (Fig.~\ref{subfig:bm_single_sim_ipos}(iii)). 
Such robustness cannot be easily obtained by extensive dynamics randomization when the policy is limited to a single jumping task (Fig.~\ref{subfig:bm_single_sim_perturb}(i), Fig.~\ref{subfig:bm_single_sim_ipos}(i)), even with additional randomization on the external perturbation (Fig.~\ref{subfig:bm_single_sim_perturb}(ii), Fig.~\ref{subfig:bm_single_sim_ipos}(ii)). 
Such a result suggests that, besides the commonly-used dynamics randomization, diversifying the tasks, like jumping to different targets or walking with different velocities, can further improve the robustness of the RL-based control policy, which is a desirable property during sim-to-real transfer.

\section{Conclusion} 
\label{sec:conclusion}
In this work, we presented an RL-based system for learning a large variety of highly-dynamic jumping maneuvers on real-world bipedal robots. 
We formulated the bipedal jumping problem as a parameterized set of tasks, and develop a goal-conditioned policy that is trained in simulation but can then be deployed directly in the real world.
In order to tackle the challenging multi-goal learning problem, we utilized a multi-stage training scheme that divides the problem into three sub-problems and addresses each through different training stages.  
We showcase that by training with multiple goals, the robot is able to generalize the learned tasks to produce robust emergent recovery behaviors from large landing impact forces or unknown perturbations.
The robustness acquired through multi-goal training then also facilitates the sim-to-real transfer process, which can not be easily acquired through single-goal training alone.  
Furthermore, we present a policy architecture that improves learning performance.
Our framework enables a real Cassie robot to perform a suite of challenging jumping tasks, such as jumping to different locations, jumping onto different evaluations, and blending multi-axes movements during a jump.  
A limitation we observe occasionally during some experiments is that the robot oscillates after a jump.
This may be due to the challenges of having a single policy for both dynamic jumps and stationary standing.
In the future, it will be interesting to combine this goal-conditioned jumping policy with a more sophisticated perception system to traverse complex environments with greater mobility.

\section*{Acknowledgments}
This work was supported in part by NSF Grant CMMI-1944722 and Canadian Institute for Advanced Research (CIFAR).
The authors would like to thank Dr. Ayush Agrawal, Xuxin Cheng, Jiaming Chen, Xiaoyu Huang, Yiming Ni, Lizhi Yang, and Bike Zhang for their gracious help.  


\bibliographystyle{plainnat}
\bibliography{references}

\pagebreak

\appendix{

\subsection{Training in Stage 1}~\label{sec:training_stage1}
\subsubsection{Reward}
The reward design in Stage 1 is presented in Table~\ref{tab:reward}.
In the first stage to initiate the training for a single jumping goal, we incentivize the robot to imitate the jumping-in-place animation. 
Therefore, the tracking rewards for motor position and foot height have overwhelming weights over others in order to accomplish a jump and stand still afterward.  
We also include the task completion term, but since the task is fixed in this stage, \textit{i.e.}, $\mathbf{c}=\mathbf{0}$, this term is more to encourage the robot to jump in place and stabilize its pelvis orientation. 
We also have a smoothing term as a small fraction of the reward at this stage and do not have the change of action reward to prevent the robot from adopting a stationary behavior at the early stage of training.

\subsubsection{Episode Design}

In the initial training stage, the episode length is designed to have $750$ timesteps corresponding to $23$ seconds.
The agent is asked to jump at $t=0$ and to stand untill the end.
If we allow the robot to stand at the beginning of the episode, the robot may focus on learning the easy standing skill and fail to explore the jumping maneuver. 
Furthermore, note that a jumping phase usually is less than $2$ second but we have a $23$-second episode. 
This is because the robot may learn jumping well but overlook the standing skill if the episode is short, which may result in the robot adopting an undesirable maneuver, such as continue hopping after landing. 
Having such a long episode can give the robot more incentive to learn a robust and stable standing skill in order to have a better return over the episode.
The early termination conditions in Stage 1 are different than the multi-goal training stage, except for the falling-over condition. 
In this stage, the foot height tracking error bound $E_e$ is smaller ($0.22$~m) while the task completion error bound $E_t$ is much larger ($[1.0, 45^\circ]$). 
This is because we want to push the robot to jump by lifting its feet at the initial stage of training and completing the task is not a big concern at this stage.

\subsection{Details of Dynamics Randomization}~\label{subsec:dynamics_rand_appendix}
The details of the dynamics parameters and randomization range used in this paper are listed in Table~\ref{tab:randomization}. 
Note that the range of the noise is relatively small (such as $0.1^\circ$ in joint position measurement and $0.5^\circ$ in joint velocity) because we found that the onboard sensors on Cassie are reliable and therefore we use a smaller bound to reduce the training complexity.  
For the robot that has larger sensor noises, a larger bound of the noise during training is recommended.

\begin{table}[ht]
\centering
\caption{Dynamics Randomization Range}
\label{tab:randomization}
\begin{tabular}{cc}
\hline
\textbf{Parameters}                    & \textbf{Range}                   \\ \hline
Floor Friction Ratio             & {[}0.3, 3.0{]}                   \\
Joint Damping                    & {[}0.3, 4.0{]} Nms/rad           \\
Spring Stiffness                 & {[}0.8, 1.2{]} $\times$ default  \\
Link Mass                        & {[}0.5, 1.5{]} $\times$ default  \\
Link Inertia                     & {[}0.7, 1.3{]} $\times$ default  \\
Pelvis (Root) CoM Position   & {[}-0.1, 0.1{]} m in $q_{x,y,z}$ \\
Other Link CoM Position      & {[}-0.05, 0.05{]} m $+$ default    \\
Motor PD Gains                   & {[}0.7, 1.3{]} $\times$ default  \\
Motor Position Noise Mean        & {[}-0.002, 0.002{]} rad          \\
Motor Velocity Noise Mean        & {[}-0.01, 0.01{]} rad/s          \\
Gyro Rotation Noise Mean         & {[}-0.002, 0.002{]} rad          \\
Linear Velocity Estimation Error & {[}-0.04, 0.04{]} m/s            \\
Communication Delay              & {[}0, 0.025{]} sec               \\ \hline
\end{tabular}
\end{table}

\subsection{Details of Baseline Models}~\label{subsec:baseline_appendix}
The details of the model structure we compared are listed as follow:
\begin{itemize}[leftmargin=9pt]
    \small
    \item \textbf{Ours} (Fig.~\ref{fig:benchmark}a): the long-term I/O history is encoded with a CNN, while the short-term I/O history is provided directly as input to the base MLP. The policy direct outputs desired motion positions. The CNN encoder and the MLP base are jointly trained.
    \item \textbf{Residual} (Fig.~\ref{fig:benchmark}b): the policy shares the same structure as the proposed one, but the policy output is a residual term added to the reference motor position at the current timestep, \textit{i.e.}, $q^d_m=\mathbf{a}_t + q^r_m(t)$, which is used in~\cite{lee2020learning,xie2020learning}. Please note that the policy has the reference motion as input. 
    \item \textbf{Long History Only} (Fig.~\ref{fig:benchmark}c): the policy only has a long-term I/O history encoded by a CNN, which is a baseline used in~\cite{lee2020learning,kumar2021rma}. Note that we still provide the robot feedback at the current timestep directly to the MLP base, as suggested by~\citet{peng2018sim}. 
    \item \textbf{Short History Only} (Fig.~\ref{fig:benchmark}d): the policy has short I/O history without the long-term I/O history CNN encoder, which is used in~\cite{li2021reinforcement} and serves as a baseline in~\cite{kumar2022adapting}. 
    \item \textbf{RMA/Teacher-Student}: an expert (teacher) policy (Fig.~\ref{fig:benchmark}e) with access to privileged environment information (listed in Table~\ref{tab:randomization}) is first trained using RL. The privileged information is encoded by an MLP into an $8$D extrinsics vector. This expert policy is then used to \textit{supervise} the training of an RMA (student) policy, which uses the base MLP copied from the expert policy, while using a long I/O history encoder to predict the teacher's extrinsic vector. This two-stage training scheme is used in~\cite{lee2020learning,kumar2021rma} and also adopted in other work such as~\cite{fu2022deep,ji2022concurrent,margolisyang2022rapid}.
    \item \textbf{A-RMA} (Fig.~\ref{fig:benchmark}g): after the standard RMA training, the parameters of the long I/O history encoder are fixed, and the base MLP is further finetuned using RL as proposed by~\citet{kumar2022adapting}. Both RMA and A-RMA are also provided with a short I/O history which are newly added in this work for a fair comparison. 
\end{itemize}

\subsection{Learning Performance in Early Stages}~\label{subsec:curve_stage12_appendix}

\begin{figure}[!]
\centering
\begin{subfigure}{0.475\linewidth}
  \centering
  \includegraphics[width=\linewidth]{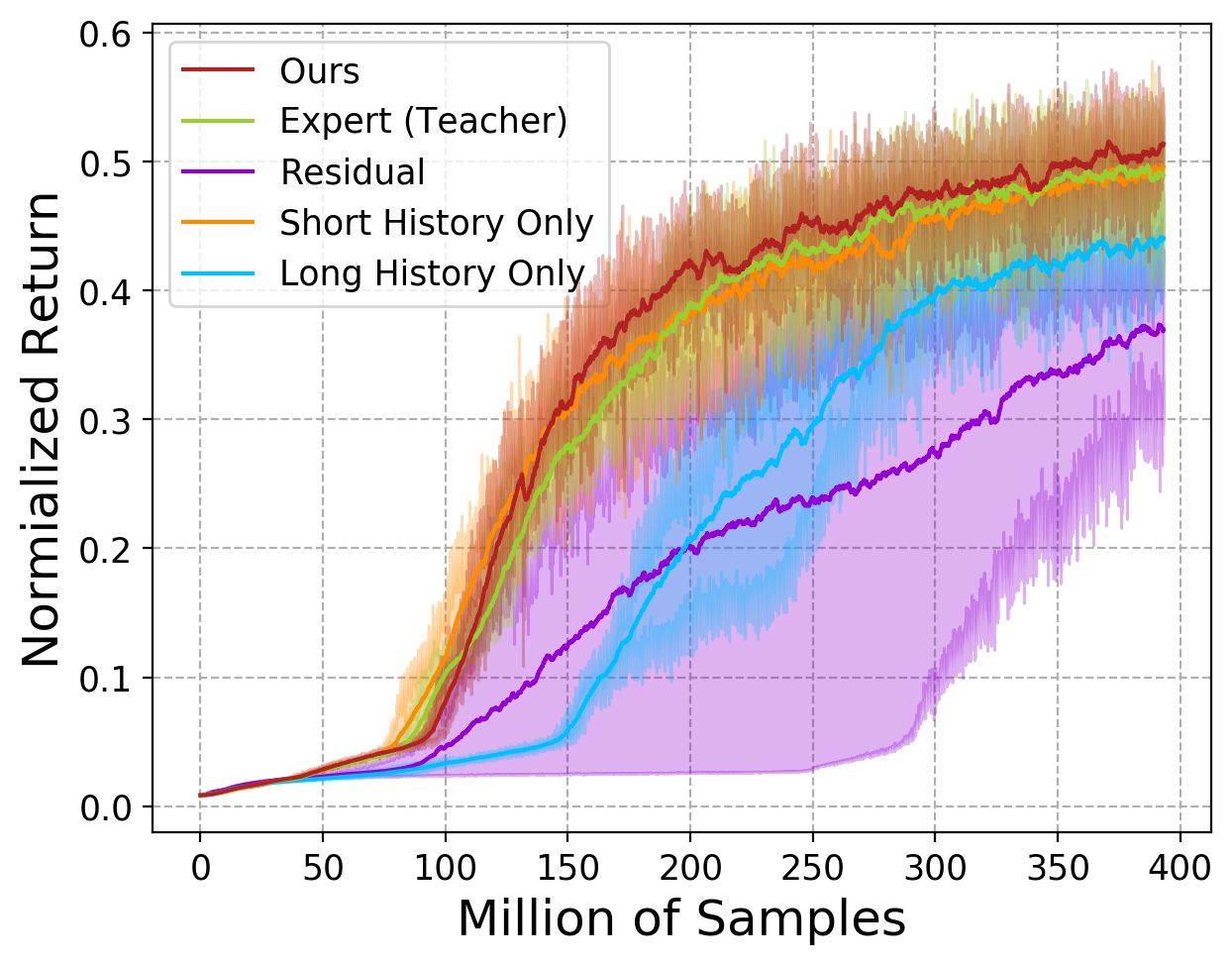}
  \caption{Stage 1: Learning a Single Goal}
  \label{subfig:stage1}
\end{subfigure}
\begin{subfigure}{0.49\linewidth}
  \centering
  \includegraphics[width=\linewidth]{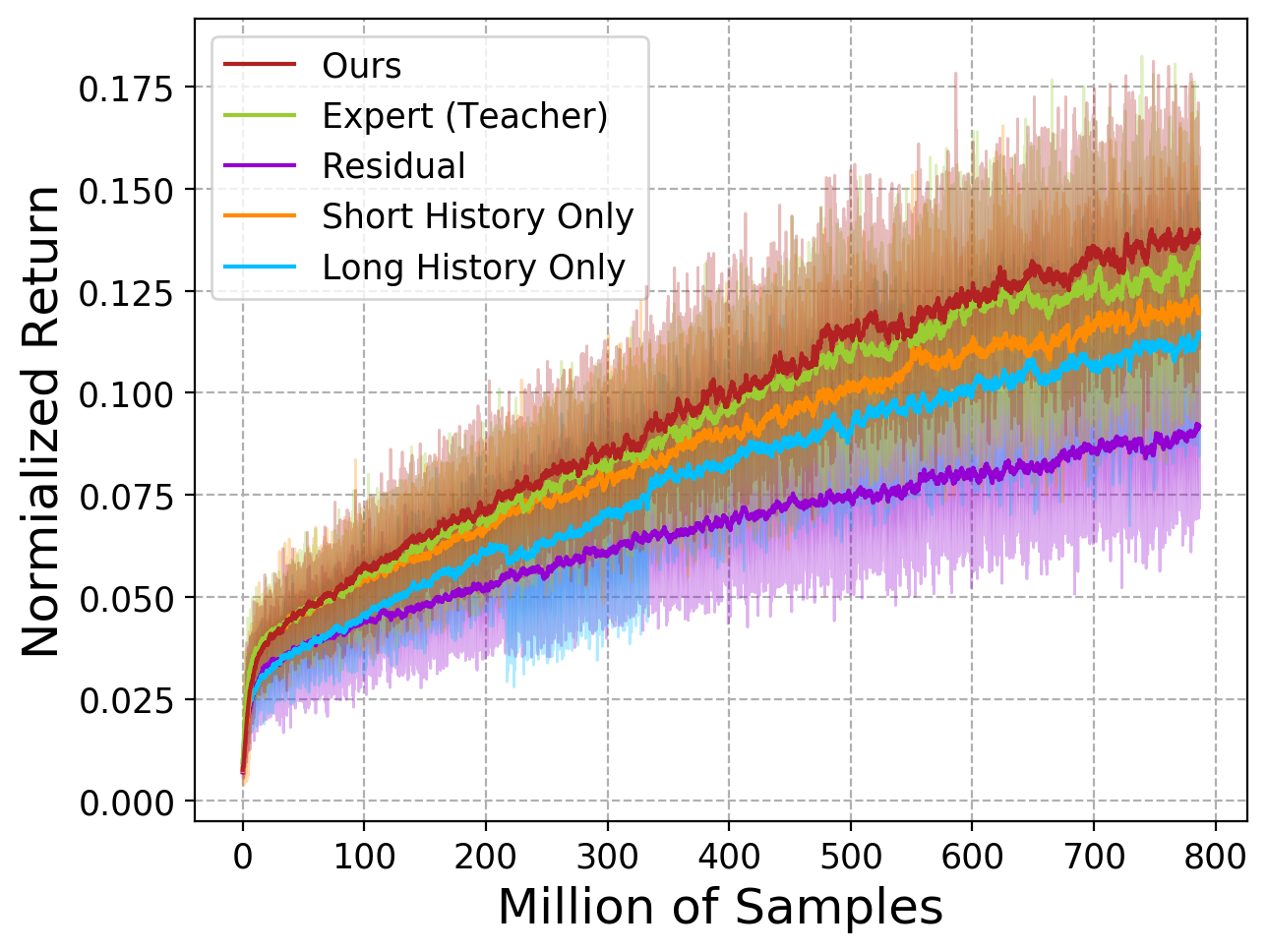}
  \caption{Stage 2: Learning Multiple Goals}
  \label{subfig:stage2}
\end{subfigure}
\caption{Benchmark of learning curves trained by different policy structures trained with $3$ random seeds in the early stages. 
The curves record the mean of normalized returns obtained using different seeds and the min and max among different seeds are the boundaries of the colored areas. 
The proposed method shows the best performance during the early stages of training including learning a single goal from scratch (Stage 1) and multiple goals in Stage 2.}
\label{fig:learning_logs_appendix}
\end{figure}

\begin{figure}[ht]
\centering
\begin{subfigure}{0.475\linewidth}
  \centering
  \includegraphics[width=\linewidth]{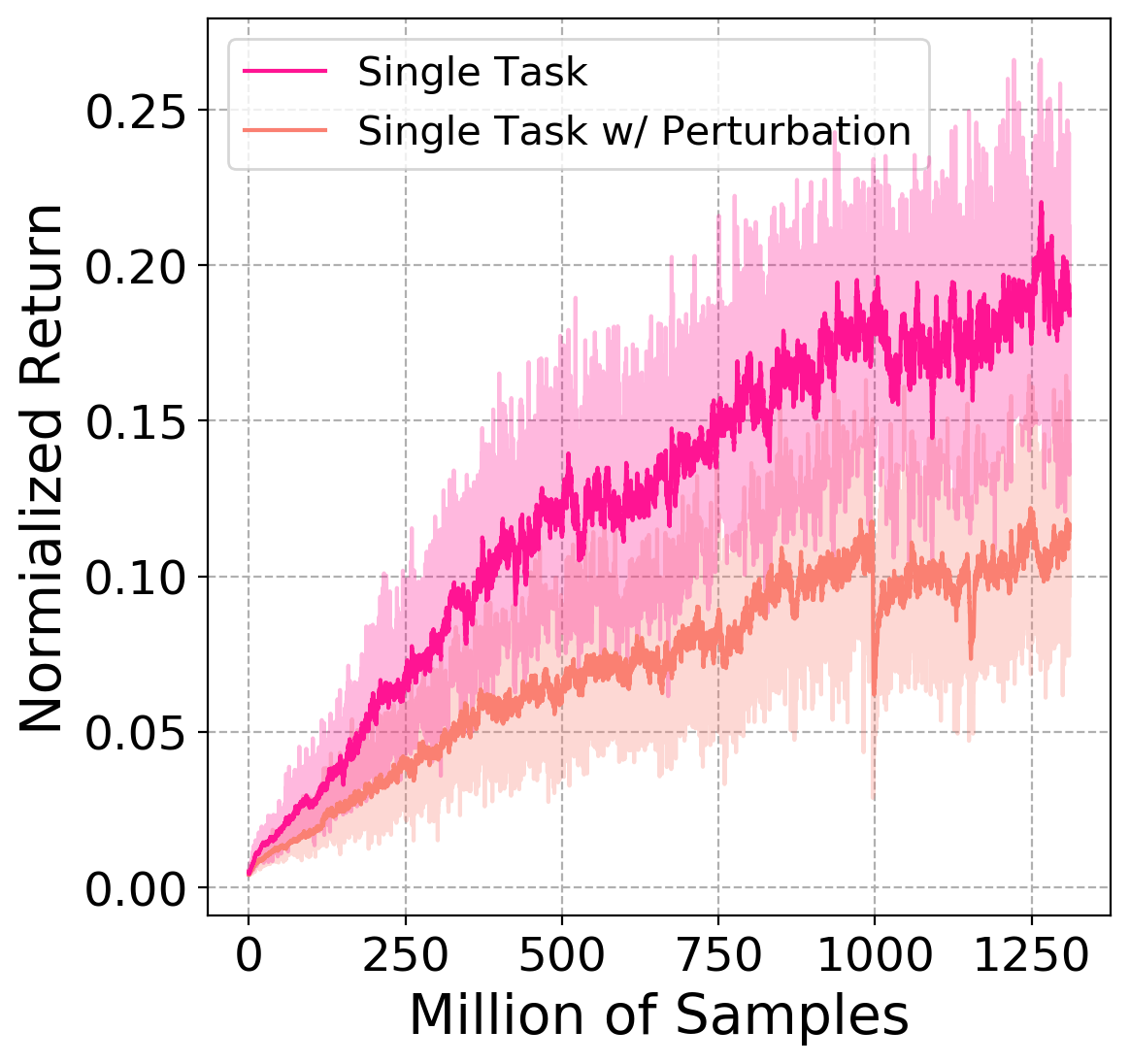}
  \caption{Learning Single Goal with Domain Randomization}
  \label{subfig:single_logs_appendix}
\end{subfigure}
\begin{subfigure}{0.475\linewidth}
  \centering
  \includegraphics[width=\linewidth]{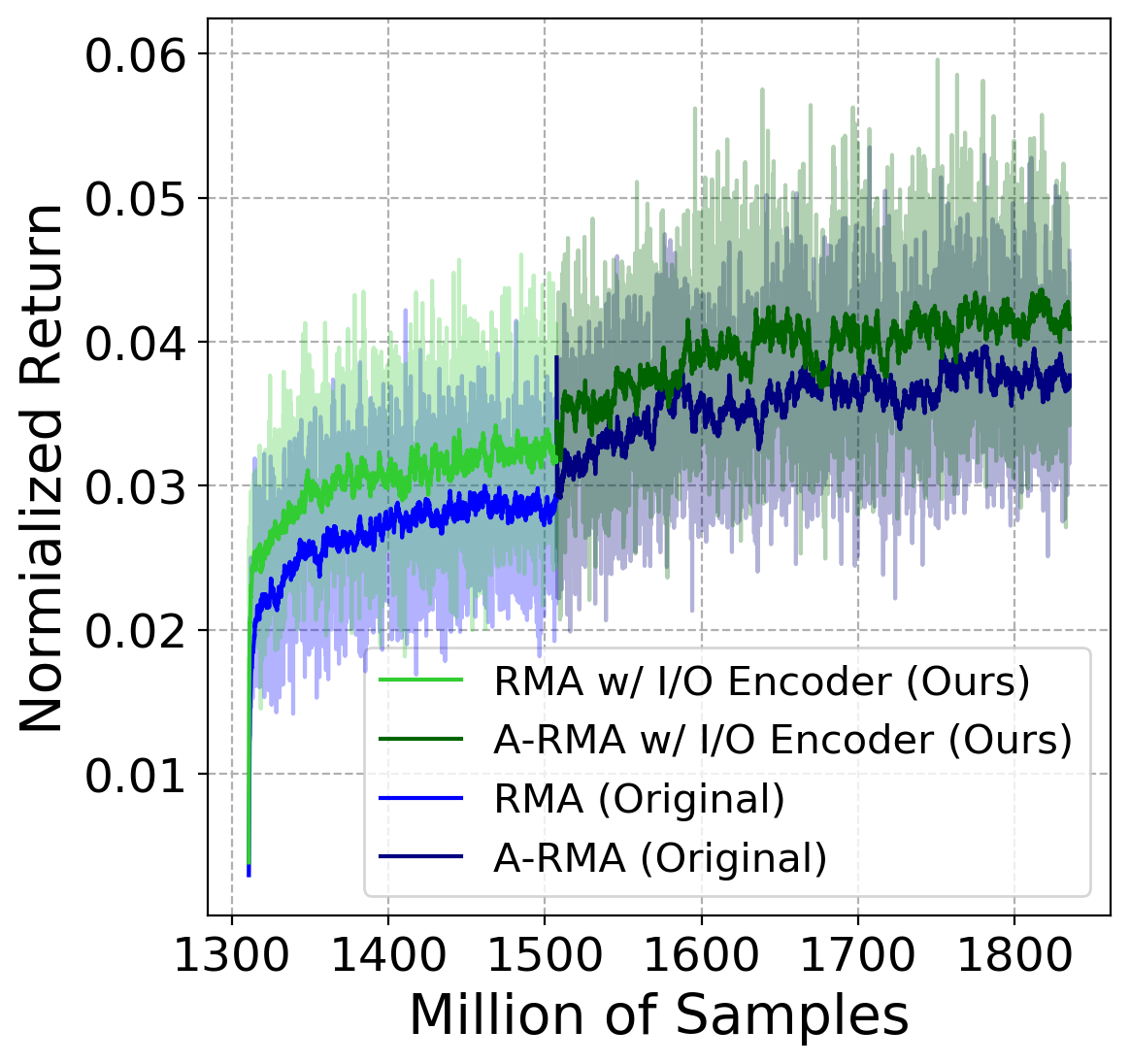}
  \caption{Learning with Different Memory Encoders for RMAs}
  \label{subfig:rma_encoders_logs_appendix}
\end{subfigure}
\caption{Additional Learning Curves.}
\label{fig:learning_curves_appendix}
\end{figure}

\begin{figure}[ht]
\centering
\includegraphics[width=\linewidth]{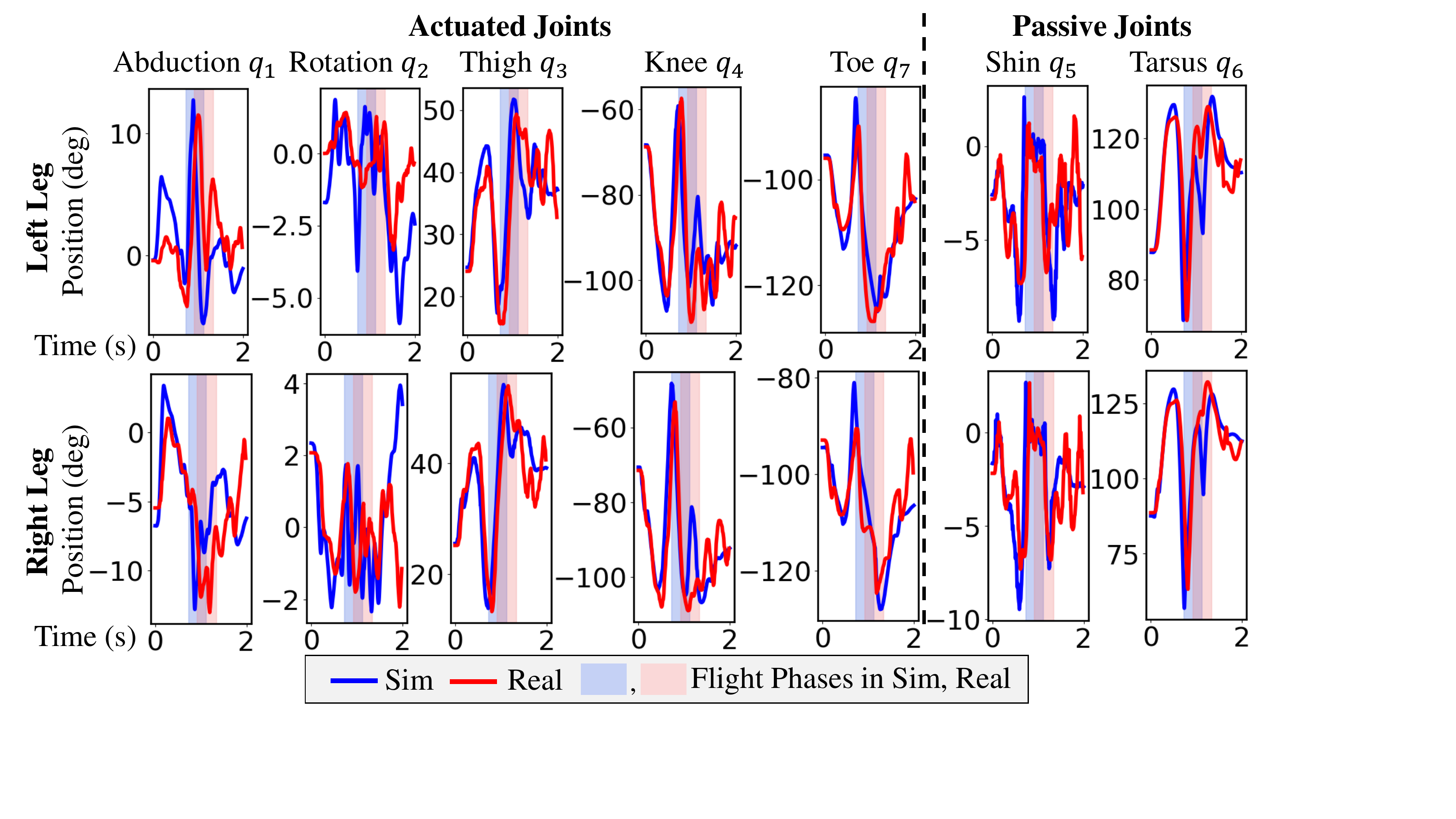}
\caption{The profiles of the robot's joint positions when it is commanded to jump to $0.44$~m-tall elevation while forward $0.88$~m in simulation and the real world, using the discrete-terrain policy.}
\label{fig:table_logs_appendix}
\end{figure}

\begin{figure}[!]
\centering
\includegraphics[width=\linewidth]{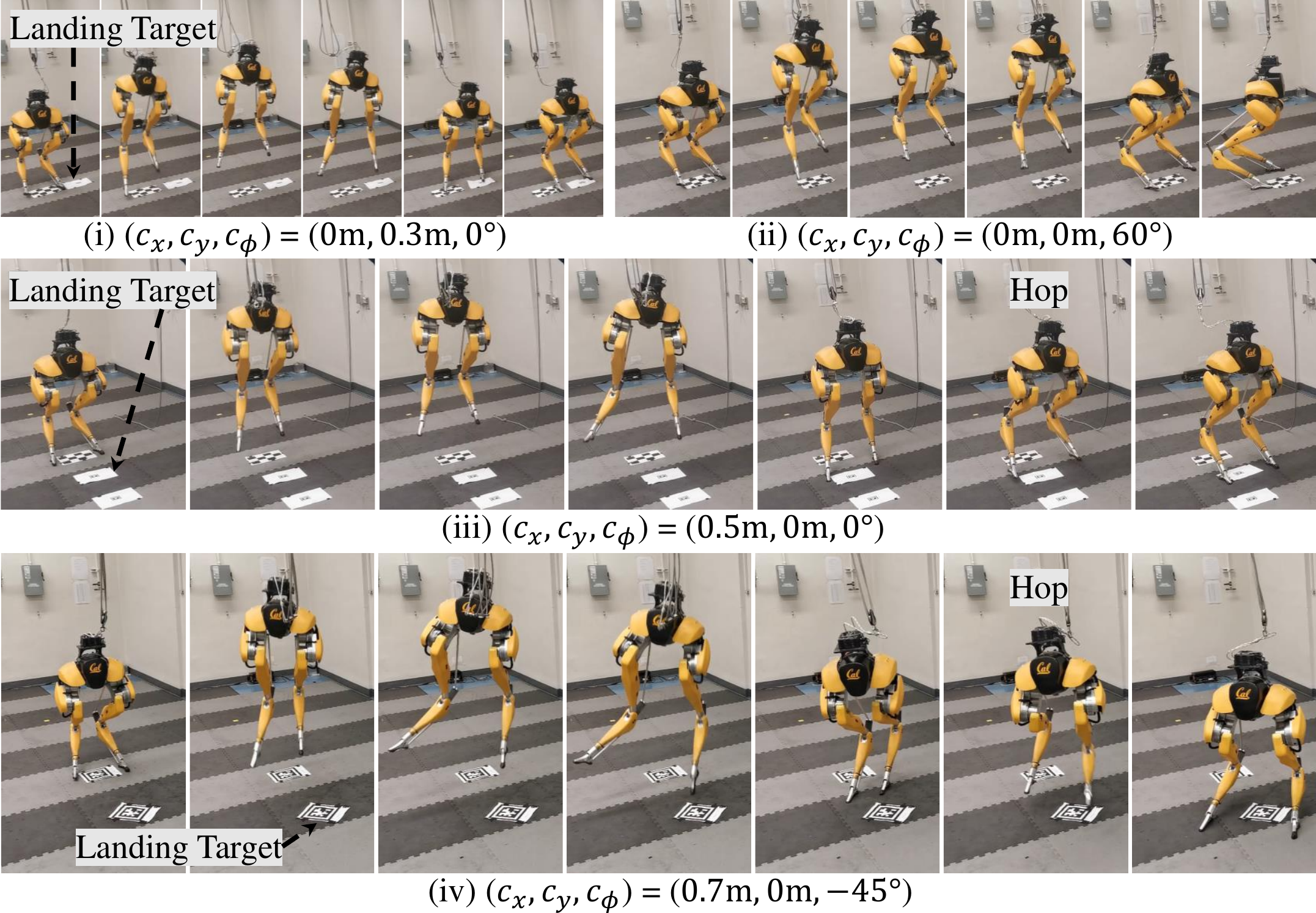}
\caption{Additional experiments show Cassie jumping to different targets with the single flat-ground policy.}
\label{fig:snapshot_flat_rest_appendix}
\end{figure}

According to Fig.~\ref{fig:learning_logs}, at the training stages without dynamics randomization (Stage 1\&2), our method shows similar, even a bit better, learning performance compared with the expert policy which has the access to the privileged environment information. 
This is because the long-term I/O history is able to provide more information than the dynamics parameters used in the expert policy, such as the robot's take-off trajectory which will be useful to determine a better landing maneuver. 
Although the policy with short history only (orange curve) shows a faster learning curve at the initial stage of training (Stage 1, Fig.~\ref{subfig:stage1}), the learning performance using short history only and long history only (blue curve) show a similar learning performance in a more complex multi-goal training stage (Stage 2).

\subsection{Additional Learning Curves}
The learning curves for single-goal policies with dynamics randomization detailed in Sec.~\ref{subsec:dynamics_randomization} is recorded in Fig.~\ref{subfig:single_logs_appendix}.
Training RMAs with different long-history encoders are recorded in Fig.~\ref{subfig:rma_encoders_logs_appendix}.
The RMA used in \cite{kumar2021rma, kumar2022adapting} (Original) has a different structure of the long-term I/O encoder (1D CNN) than the one used in this work. It has 3 hidden layers and the [kernel size, filter size, stride size] of each layer is $[8, 32, 4]$, $[5, 32, 1]$, and $[5, 32, 1]$, with zero padding, respectively.

\subsection{Additional Hardware Experiments}

More experiment results are presented in Fig.~\ref{fig:table_logs_appendix} and Fig.~\ref{fig:snapshot_flat_rest_appendix}. 
It shows the capacity of the flat-ground policy to accomplish more challenging jumping tasks on the real robot Cassie.

}

\end{document}